\documentclass[10pt,journal,compsoc]{IEEEtran}
\usepackage[nocompress]{cite}
\usepackage[pdftex]{graphicx}
\usepackage{amsmath}
\usepackage{algorithmic}
\usepackage{fixltx2e}
\usepackage{stfloats}
\usepackage{comment}
\ifCLASSOPTIONcaptionsoff
  \usepackage[nomarkers]{endfloat}
 \let\MYoriglatexcaption\caption
 \renewcommand{\caption}[2][\relax]{\MYoriglatexcaption[#2]{#2}}
\fi
\usepackage{url}

\usepackage{amssymb}
\usepackage{amsthm}
\usepackage{algorithm}
\usepackage{array}
\usepackage{relsize}
\usepackage{wrapfig}
\usepackage{mathtools}
\usepackage{bm}
\usepackage{calc} 
\usepackage{enumitem}
\usepackage[labelformat=simple]{subcaption}
\usepackage{xcolor}
\usepackage{pdfpages}

\DeclareMathOperator*{\argmax}{argmax}
\DeclareMathOperator*{\argmin}{argmin}

\newtheorem{lemma}{Lemma}[section]

\newtheorem{proposition}{Proposition}[section]

\newcommand{\eg}{e.g.}
\newcommand{\ie}{i.e.}

\newcommand{\etc}{etc.}

\begin{document}
\title{On Connections between Regularizations for Improving DNN Robustness}

\author{Yiwen Guo, Long Chen, Yurong Chen, and Changshui Zhang,~\IEEEmembership{Fellow,~IEEE}
 \IEEEcompsocitemizethanks{
  \IEEEcompsocthanksitem Y. Guo is with Bytedance AI Lab. E-mail: guoyiwen.ai@bytedance.com.
  \IEEEcompsocthanksitem L. Chen is with the Academy for Advanced Interdisciplinary Studies, Center for Data Science, Peking University, Beijing 100871, China. E-mail: xidonglc@gmail.com. 
  \IEEEcompsocthanksitem Y. Chen is with Intel Labs China. E-mail: yurong.chen@intel.com.
  \IEEEcompsocthanksitem C. Zhang is with the Institute for Artificial Intelligence, Tsinghua University (THUAI), the State Key Lab of Intelligent Technologies and Systems, Beijing National Research Center for Information Science and Technology (BNRist), the Department of Automation, Tsinghua University, 
Beijing, 100084, China. E-mail: zcs@mail.tsinghua.edu.cn.}
 \thanks{Y. Guo and L. Chen contribute equally to this work.}}



\IEEEtitleabstractindextext{
 \begin{abstract}
This paper analyzes regularization terms proposed recently for improving the adversarial robustness of deep neural networks (DNNs), from a theoretical point of view.
Specifically, we study possible connections between several effective methods, including input-gradient regularization, Jacobian regularization, curvature regularization, and a cross-Lipschitz functional.
We investigate them on DNNs with general rectified linear activations, which constitute one of the most prevalent families of models for image classification and a host of other machine learning applications.
We shed light on essential ingredients of these regularizations and re-interpret their functionality.
Through the lens of our study, more principled and efficient regularizations can possibly be invented in the near future.
 \end{abstract}

 \begin{IEEEkeywords}
Deep neural networks, adversarial robustness, regularizations, network property
 \end{IEEEkeywords}}

\maketitle
\IEEEdisplaynontitleabstractindextext
\IEEEpeerreviewmaketitle

\IEEEraisesectionheading{\section{Introduction}\label{sec:intro}}
\IEEEPARstart{I}{t} has been discovered that deep neural networks (DNNs) are vulnerable to adversarial examples~\cite{Szegedy2014, Goodfellow2015, Madry2018}, and the phenomenon can prohibit them from being deployed in security-sensitive applications. 
Amongst the most effective methods for mitigating the issue, \emph{adversarial training}~\cite{Szegedy2014,Goodfellow2015,Madry2018} is capable of resisting a series of malicious examples~\cite{Madry2018,Tramer2018} and yield adversarially robust DNN models in the sense of an $l_p$ norm. 
By injecting advanced adversarial examples (\eg, using BIM~\cite{Kurakin2017} or PGD~\cite{Madry2018}) into training as some sort of augmentation, the obtained models learn to defend against these examples.
In addition, the obtained models may also resist some other types of adversarial examples (generated using, for example, the fast gradient sign method~\cite{Goodfellow2015}).  
However, advanced adversarial examples are typically generated in an iterative manner by back-propagating deep models for multiple times, and thus the mechanism may demand a massive amount of computation~\cite{Moosavi2019}.

Another thriving category of methods for hardening DNNs is to perform \emph{regularizations}, aim at trading off the effectiveness and efficiency properly. 
Although most traditional regularization-based strategies (\eg, weight decay~\cite{Krogh1992} and dropout~\cite{Srivastava2014}) do \emph{not} operate properly in this respect, a variety of recent work~\cite{Cisse2017, Hein2017,Ross2018,Jakubovitz2018,Moosavi2019} has shown that more dedicated and principled regularizations help to gain comparable or only slightly worse performance in improving DNN robustness. 
Instead of raising a perpetual  ``arms-race'', these regularization-based strategies are in general attack-agnostic and of benefit to the generalization ability~\cite{Cisse2017} and interpretability of learning models~\cite{Ross2018}.  
Moreover, the computational and memory complexity of these methods are acceptable in very large models. 
It has also been shown that the methods can be combined with adversarial training to achieve even stronger DNN robustness.

While many regularizers have been developed for DNN robustness, there is of yet few comparative analysis among these choices, especially from a theoretical point of view.
In this paper, we attempt to shed light on intrinsic functionality and theoretical connections between several effective regularizers, even if their formulations may stem from different rationales.  
Concretely, it has been presented over the past few years that regularizing the Euclidean norm of an input-gradient~\cite{Ross2018, Lyu2015}, the Frobenius norm of a Jacobian matrix~\cite{Jakubovitz2018, Sokolic2017}, the spectral norm of a Hessian matrix~\cite{Moosavi2019}, and a cross-Lipschitz functional~\cite{Hein2017} all significantly contribute to the adversarial robustness of DNNs. 
We analyze all these choices on DNNs with general rectified linear activations, which are ubiquitous in image classification and a host of other machine learning tasks.

Some of our key contributions and observations are:

\begin{description}[leftmargin=!,labelwidth=\widthof{\bfseries $\bullet\quad$},font=$\bullet$]
\item [\quad] We present, for the first time, an analytic expression for the $l_2$ norm of an approximately-optimal adversarial perturbation concerned in very recent papers~\cite{Moosavi2019, Simon2018}, to demonstrate that local cross-Lipschitz constants~\cite{Hein2017} and the prediction probability are its essential ingredients in binary classification cases. 
In addition to the $l_2$ norm-based results, we also show similar results for the robustness to $l_\infty$ norm-based attacks.

\item [\quad] We unveil that most discussed regularizations advocate small local cross-Lipschitz constants in binary classification, except for the Jacobian regularization that suggests small local Lipschitz constants, yet regularizing the two network properties can be equivalent.

\item [\quad] We further demonstrate that critical discrepancies still exist between specific methods, mostly in regularizing the prediction probability/confidence.

\item [\quad] We extend some analyses to multi-class classification and verify our findings with experiments.
\end{description}

\section{Regularizations Improving Robustness}


\subsection{Adversarial Phenomenon in DNNs}
Given an input instance $\mathbf x\in\mathbb R^n$, a DNN-based classifier offers its prediction along with a softmax normalized probability $p(\mathbf x)_k=\exp(\mathbf z_k)/\sum\exp(\mathbf z_j)$ for each class $k$ on top of a vector representation $\mathbf z= g(\mathbf x)$. 
Suppose that a set of labeled instances $\{(\mathbf x_i, y_i)\}_i$ are provided, then a classifier is typically learned with the assistance of an objective function $\mathcal L(\cdot, \cdot)$ that evaluates training prediction loss, \ie, the average discrepancy between a set of predictions $\{p(\mathbf x_i)\}_i$ and ground-truth $\{y_i\}_i$. 

Existing adversarial attacks can be roughly divided into two main categories, \ie, white-box attacks~\cite{Szegedy2014,Goodfellow2015} and black-box attacks~\cite{Papernot2017,Chen2017}, according to how much information of the victim model is accessible to an adversary~\cite{Carlini2017}.
Our study in this paper mainly focuses on the white-box non-targeted attacks and defenses against them, in order to be complied with prior theoretical work. 
Under such threat, substantial endeavors have been exerted to demonstrate the adversarial vulnerability of DNNs~\cite{Szegedy2014,Goodfellow2015,Papernot2016,Moosavi2016,Carlini2017,Chen2018,Athalye2018,Madry2018}.
Most of them are proposed within a framework that favors perturbations with least $l_p$ norms yet would still cause the DNNs to make incorrect predictions.
That being said, an adversary opts to solve
\begin{equation}\label{eq:1}
\min_{\mathbf r} \|\mathbf r\|_p\quad \textrm{s.t.}\ \argmax_k g(\mathbf x+\mathbf r)_k \neq\argmax_k g(\mathbf x)_k.
\end{equation}
Utilizing the objective function $\mathcal L(\cdot, \cdot)$, the task of mounting adversarial attacks can be formulated from a dual perspective which attempts to maximize the loss with a presumed perturbation magnitude (in the context of $l_p$ norms). That being said, given $\epsilon>0$, one may resort to
\begin{equation}\label{eq:2}
\max_{\|\mathbf r\|_p\leq \epsilon} \mathcal L(\mathbf x+\mathbf r, y). 
\end{equation}
Omit box constraints on the image domain, many off-the-shelf attacks~\cite{Goodfellow2015,Papernot2016,Moosavi2016,Carlini2017,Madry2018} can be considered as efficient approximations to either~(\ref{eq:1}) or~(\ref{eq:2}). 
Under certain circumstances, their solutions can be equivalent to the optimal solutions to~(\ref{eq:1}) or~(\ref{eq:2}).
For instance, the fast gradient sign method (FGSM)~\cite{Goodfellow2015} achieves the optimum of~(\ref{eq:2}) with some binary linear classifiers and $p=\infty$~\cite{Guo2018}. 
Also, take any linear model together with $p=2$, the DeepFool perturbation~\cite{Moosavi2016} is theoretically optimal to~(\ref{eq:1}). 
Training with an augmented set involving adversarial examples, \ie, adversarial training, has been proven to be very effective in improving the DNN robustness~\cite{Madry2018}, regardless of the computational burden. 

\subsection{Regularizations and Important Notations}
A recent study~\cite{Simon2018} demonstrates the relationship between a classical regularization~\cite{Drucker1991} and adversarial training~\cite{Goodfellow2015}.
It is conceivable that a principled regularization term involved in training suffices to yield DNN models with comparable robustness, whereby a whole series of methods have been developed. 
Unlike many traditional methods which are normally date-independent (\eg, weight decay and dropout), recent progress conforms closely with theoretical guarantees and focuses mostly on regularizing the loss landscape~\cite{Lyu2015, Sokolic2017, Hein2017,Ross2018,Jakubovitz2018,Moosavi2019}. 
Before systematically studying their functionality and relationships in the following sections, we first introduce some important notations.

Given the objective function $\mathcal L(\cdot, \cdot)$ for classification, we will refer to 1) $\nabla:=\nabla_{\mathbf x} \mathcal L(\mathbf x, y)$, as its gradient with respect to (w.r.t.) the input vector $\mathbf x$, 2) $H$, as the Hessian matrix of $\mathcal{L}$, and 3) $J$, as the Jacobian matrix of $g(\mathbf x)$ w.r.t. $\mathbf x$. 
It has been presented that training regularized using $\|J\|_F$---the Frobenius norm of $J$ (dubbed the \emph{Jacobian regularization},~\cite{Jakubovitz2018, Sokolic2017}), $\|\nabla\|_2$---the Euclidean norm of $\nabla$ (\ie, the \emph{input-gradient regularization},~\cite{Ross2018, Lyu2015}), $\|H\|_2$---the spectral norm of $H$ (\ie, the \emph{curvature regularization},~\cite{Moosavi2019}), and a cross-Lipschitz functional~\cite{Hein2017} as will be elaborated later all significantly improve the adversarial robustness of obtained models.
We focus on DNNs with general rectified linear units (general ReLUs)~\cite{Nair2010,He2015,Shang2016} as nonlinear activations and analyze in binary classification and multi-class classification tasks separately in the following sections.

\section{Binary Classification}\label{sec:bc}

With the background information introduced in the previous section, here we first discuss different regularizations in binary classification DNNs, and we will generalize some of our results to multi-class classifications in Section~\ref{sec:mc}. 

For simplicity of notations, let us first consider a multi-layer perceptron (MLP) parameterized by a series of weight matrices $W_1\in \mathbb R^{n_0\times n_1},\ldots, W_d\in \mathbb R^{n_{d-1}\times n_{d}}$, where $n_0=n$ and $n_d=2$ in our theories. \emph{(We stress that, although a simple MLP is formulated here, our following discussions directly generalize to DNNs with convolutions, poolings, skip-connections~\cite{He2016}, self-attentions~\cite{Bahdanau2015}, \etc})\ For a $d$-layer MLP, we have
\begin{equation}
g(\mathbf x) = W^T_d\sigma(W^T_{d-1}\sigma(\ldots\sigma(W^T_1\mathbf x))),
\end{equation} 
in which the general ReLU activation $\sigma(\cdot)$ of our particular interest is piecewise linear and hence $g(\cdot)$ is also piecewise linear. 
Following prior work~\cite{Guo2018}, we can define $\mathbf a_0 := \mathbf x$ and $\mathbf a_j:=\sigma(W^T_j\mathbf a_{j-1})=D^T_j(\mathbf x)W^T_j\mathbf a_{j-1}$, for $1\leq j\leq d$, in which 
\begin{equation}
D_j(\mathbf x) := \mathrm{diag}\left(1_{W_j[:,1]^T \mathbf a_{j-1} > 0},\ldots,1_{W_j[:,n_{j}]^T \mathbf a_{j-1} > 0}\right)
\end{equation}
is an $n_j\times n_j$ diagonal matrix whose main diagonal entries corresponding to nonzero activations within the $j$-th parameterized layer take an value of $+1$, and others take an value of $0$.
Denote by $\mathbf w_{\pm}$ the two columns of matrix $W_d$ (\ie, $W_d=[\mathbf w_{+},\mathbf w_{-}]$), we have the two entries of $p(\mathbf x)$ as: $p(\mathbf x)_{+}=\exp(\mathbf w_{+}^T \mathbf a_{d-1})/\sum\exp(\mathbf w_{\pm}^T \mathbf a_{d-1})$ and $p(\mathbf x)_{-}=1-p(\mathbf x)_{+}$. 
These two scalars estimate the probability of $\mathbf x$ being sampled from the positive and negative classes, respectively.
Since $g(\cdot)$ is piecewise linear as analyzed, there exists a polytope $Q({\mathbf x})$ to which the input instance $\mathbf x$ belongs and on which $g(\cdot)$ is linear, \ie, $D_j(\mathbf x')=D_j(\mathbf x)$ and 
\begin{equation}\label{eq:5}
\left.g(\mathbf x')\right|_{\mathbf x' \in Q({\mathbf x})} = V^T\mathbf x', 
\end{equation}
in which $V=[\mathbf v_{+},\mathbf v_{-}]$ is a matrix with its columns $\mathbf v_{\pm}:= W_1 D_1(\mathbf x)\ldots W_{d-1} D_{d-1}(\mathbf x)\mathbf w_{\pm}$.

\subsection{Robustness in Binary Classification}\label{sec:binary_rob}
Our analyses stem from Problem~(\ref{eq:1}). 
For binary classification with $y\in\{+1, -1\}$, we can rewrite the optimization problem as: $\min_{\mathbf r} \|\mathbf r\|_p\ \textrm{s.t.}\ \mathcal L(\mathbf x+\mathbf r, y) \geq \beta$, just as suggested~\cite{Moosavi2019}, in which $\beta$ is a threshold for correct and incorrect classifications and its value solely depends on the choice of the loss function $\mathcal L(\cdot, \cdot)$ (\eg, if the cross-entropy loss is chosen, then $\beta=\log(2)$).
It follows from DeepFool and others~\cite{Moosavi2019} that we may well-approximate the constraint with a Taylor series and get bounds for the ($l_2$) magnitude of $\mathbf r^*:=\argmin \|\mathbf r\|_2\ \textrm{s.t.}\ \mathcal L(\mathbf x, y)+\nabla^T\mathbf r+\mathbf r^TH\mathbf r/2 \geq \beta$, as will be presented in Lemma~\ref{lem:1} as below. 

\begin{lemma}\textnormal{\cite{Moosavi2019}}\label{lem:1}
Let $\mathbf x$ be a correctly classified instance such that $\xi:=\beta-\mathcal L(\mathbf x, y)\geq 0$, and let $\mathbf u\in \mathbb R^n$ be the normalized eigenvector corresponding to the largest eigenvalue of $H$, then we have  
\begin{equation}\label{eq:6}
\begin{aligned}
\frac{\|\nabla\|_2}{\|H\|_2}\left(\sqrt{1+\frac{2\|H\|_2\xi}{\|\nabla\|^2_2}}-1\right) \leq \|\mathbf r^*\|_2 \leq \\
 \frac{|\nabla^T\mathbf u|}{\|H\|_2}\left(\sqrt{1+\frac{2\|H\|_2\xi}{|\nabla^T\mathbf u|^2}}-1\right).
\end{aligned}
\end{equation}
\end{lemma}

The above lemma establishes connections between the robustness of a DNN and the spectral norm of its Hessian matrix $H$.
Though enlightening, the variables $\mathbf u$, $\|\nabla\|_2$, and $\|H\|_2$ in Eq.~(\ref{eq:6}) are heavily entangled so that it is difficult to reveal the functionality of concerned regularizations. 

Fortunately, we show that the derived bounds are tight such that they collapse to the same expression in terms of $p(\mathbf x)_y$ and a local cross-Lipschitz constant~\cite{Hein2017} in binary classification with some common choices of the loss function (\eg, the cross-entropy loss and logistic loss). 
To be concrete, suppose that the cross-entropy loss is adopted, then with the $n\times 2$ matrix $V=[\mathbf v_{+},\mathbf v_{-}]$ introduced in Eq.~(\ref{eq:5}), we have the following lemma and theorem.

\begin{lemma}\textnormal{(Simplified expressions for $J$, $\nabla$, and $H$).}\label{lem:2}
Given an instance paired with its label $(\mathbf x, y)$, we have for the Jacobian $J$, input-gradient $\nabla$, and Hessian $H$:
\begin{equation}\label{eq:7}
\begin{aligned}
J & = V,\\
\nabla &= y(p(\mathbf x)_y-1) (\mathbf v_{+} -\mathbf v_{-})^T, \\
H & = p(\mathbf x)_{+}p(\mathbf x)_{-} (\mathbf v_{+}-\mathbf v_{-}) (\mathbf v_{+}-\mathbf v_{-})^T \\
& = p(\mathbf x)_y(1-p(\mathbf x)_y) (\mathbf v_{+}-\mathbf v_{-}) (\mathbf v_{+}-\mathbf v_{-})^T.
\end{aligned}
\end{equation}
\end{lemma}

\begin{proposition}\textnormal{(An analytic expression for $\|\mathbf r^*\|_2$).}\label{pro:1}
For the binary classifier with a locally linear $g(\cdot)$ and a correctly classified instance $\mathbf x$, we have  
\begin{equation}\label{eq:8}
\|\mathbf r^*\|_2 =  \frac{1}{p(\mathbf x)_y\|\mathbf v_{+}-\mathbf v_{-}\|_2}\left(\sqrt{1+\frac{2p(\mathbf x)_y\xi}{1-p(\mathbf x)_y}} - 1\right).
\end{equation}
\end{proposition}

\begin{figure}[t]
	\centering 
	\begin{subfigure}[b]{0.48\linewidth}
		\includegraphics[trim={0.0in 0.0in 0.4in 0.7in},clip, width=\linewidth]{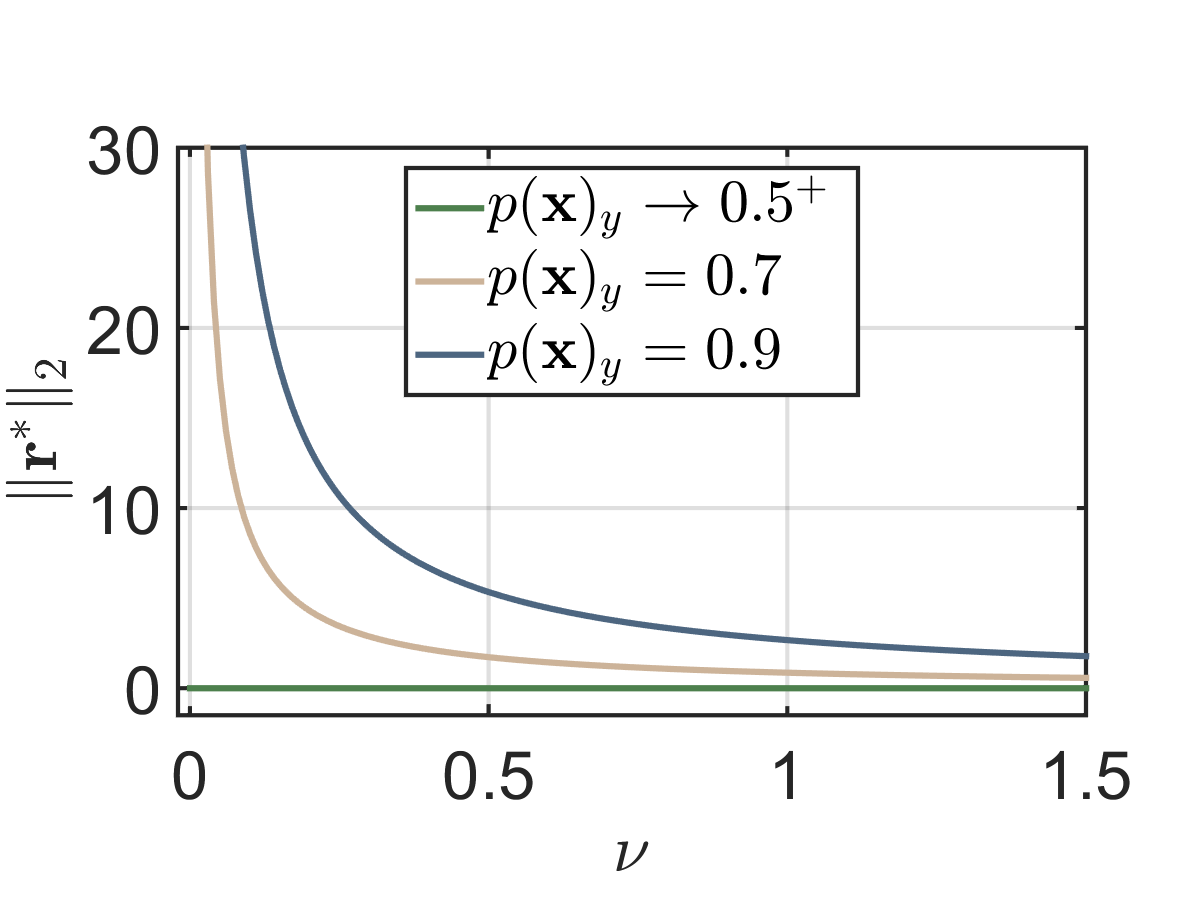}	
	\end{subfigure} \hskip 0.05in
	\begin{subfigure}[b]{0.48\linewidth}
		\includegraphics[trim={0.0in 0.12in 0.7in 0.8in},clip, width=\linewidth]{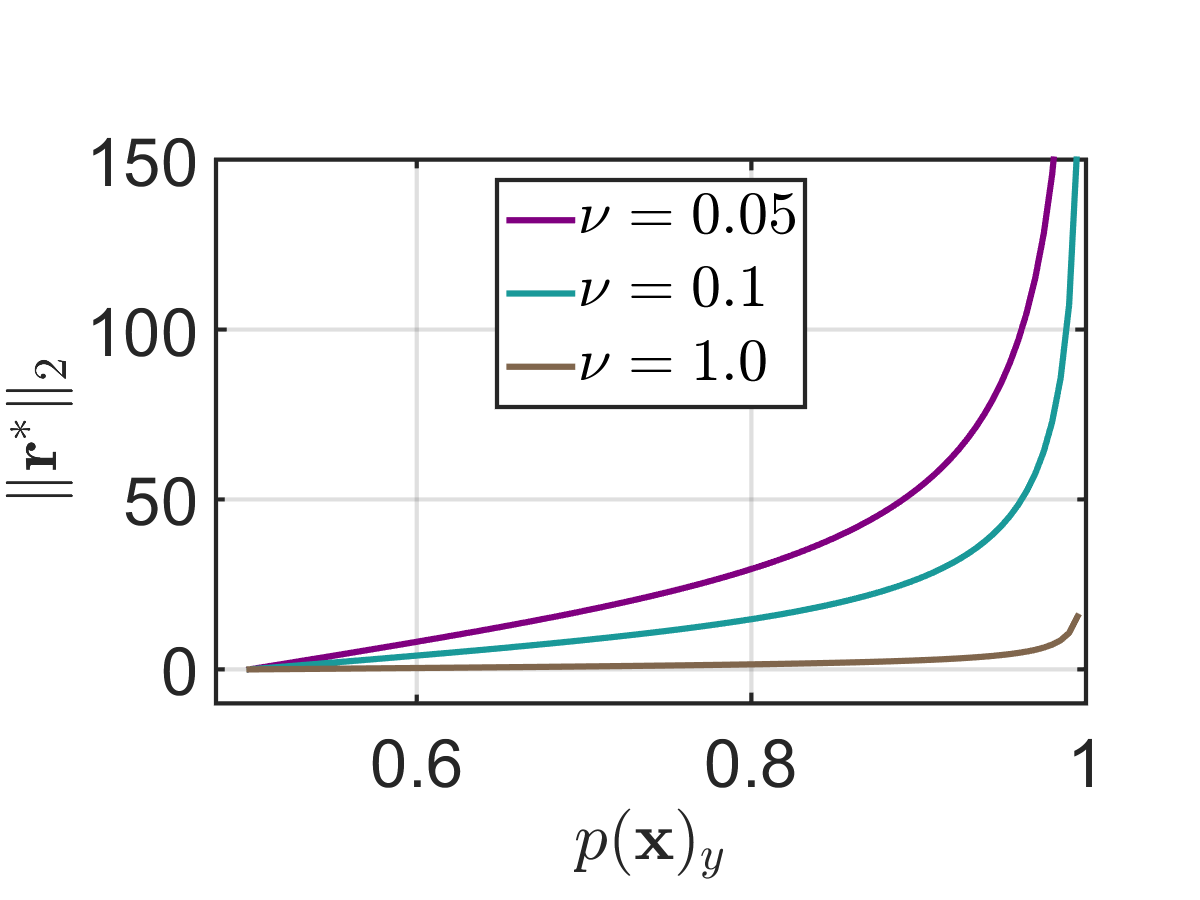}
	\end{subfigure}
	
	\caption{Illustration of how $\|\mathbf r^\ast\|_2$ varies with the prediction probability $p(\mathbf x)_y$ in binary scenarios.}\vskip -0.1in
\label{fig:1}
\end{figure}

Proposition~\ref{pro:1} is obtained on the basis of Lemma~\ref{lem:1} and~\ref{lem:2}. 
See our proofs in Appendix A and B, respectively. 
Similar results can be achieved with the logistic loss (as also demonstrated in the appendix).
The decomposition of $\|\mathbf r^*\|_2$ (\ie, the $l_2$ magnitude of $\mathbf r^\ast$) in the derived Eq.~(\ref{eq:8}) appears to be more obvious than in Eq.~(\ref{eq:6}), and it can be concluded that $\xi$, $p(\mathbf x)_y$, and $\|\mathbf v_{+} - \mathbf v_{-}\|_2$ jointly affect the $l_2$ magnitude of $\mathbf r^\ast$. 
\emph{Seeing that the value of $\xi$ is determinate w.r.t. $p(\mathbf x)_y$, the prediction probability $p(\mathbf x)_y$ and $\|\mathbf v_{+} - \mathbf v_{-}\|_2$ become the only dominating ingredients. }
Let us define $\nu:=\|\mathbf v_{+} - \mathbf v_{-}\|_2$ which is in fact a local cross-Lipschitz constant of $g(\cdot)$~\cite{Hein2017} for better clarity.
Even though $\nu$ might as well be influential to the prediction probability $p(\mathbf x)_y$, we discuss them separately here, considering that the latter can still be optimized with any presumed value of the former~\footnote{Within linear networks where all instances share the same (data-independent) $\mathbf v_{+}$ and $\mathbf v_{-}$, we can still have different prediction probabilities for different input instances. }.

It is easy to verify that $\|\mathbf r^*\|_2=0$ holds for all $\nu>0$, in a special case of $p(\mathbf x)_y\rightarrow 0.5^{+}$. Yet, for $p(\mathbf x)_y>0.5$, the general impact of the prediction probability $p(\mathbf x)_y$ in Eq.~(\ref{eq:8}) is still obscure.
To gain direct insights, we depict how $\|\mathbf r^*\|_2$ varies with $p(\mathbf x)_y\in(0.5, 1.0]$ on the right panel of Figure~\ref{fig:1}, given specific $\nu$ values. 
We observe that, in general, a larger $p(\mathbf x)_y$ implies a larger $\|\mathbf r^*\|_2$ and thus lower vulnerability of a classification model, provided that the $l_2$ magnitude of $\mathbf r^*$ is a reasonable measure of the robustness and $p(\mathbf x)_y>(1-p(\mathbf x)_y)$ (or equivalently, $p(\mathbf x)_y>0.5$). 
See also the left panel of the figure for an illustration with $p(\mathbf x)_y$ approaching $0.5$ from above, being equal to $0.7$, and being equal to $0.9$.

Our theoretical result in Proposition~\ref{pro:1} gives rise to a formal guarantee of the $l_2$ robustness for piecewise linear DNNs, without concerning much about the accuracy of the Taylor approximation.
Regarding the adversarial robustness to some other $l_p$ norm-based attacks, we have similar results in this paper.
One might be of special interest to the $p=\infty$ case as it has been widely considered in practical attacks.
Proposition~\ref{pro:1.2} and~\ref{pro:1.1} provide results from different viewpoints in correspondence to~(\ref{eq:2}) and (\ref{eq:1}), \ie, by bounding the worst-case loss $\mathcal \eta^\ast:=\max_{\|\mathbf r\|_\infty\leq \epsilon}\mathcal L(\mathbf x, y)+\nabla^T\mathbf r+\mathbf r^TH\mathbf r/2$ with any fixed $\epsilon>0$ and by providing an analytic expression for the $l_\infty$ norm of $\tilde {\mathbf r}^\ast:=\argmin \|\mathbf r\|_\infty\ \textrm{s.t.}\ \mathcal L(\mathbf x, y)+\nabla^T\mathbf r+\mathbf r^TH\mathbf r/2 \geq \beta$.~\footnote{Note that we probably have $\mathbf r^\ast \neq\tilde{\mathbf r}^\ast$.}

\begin{figure*}[t]
	\centering 
	\begin{subfigure}[b]{0.241\linewidth}
		\includegraphics[trim={0.0in 0.05in 0.5in 0.3in},clip, width=\linewidth]{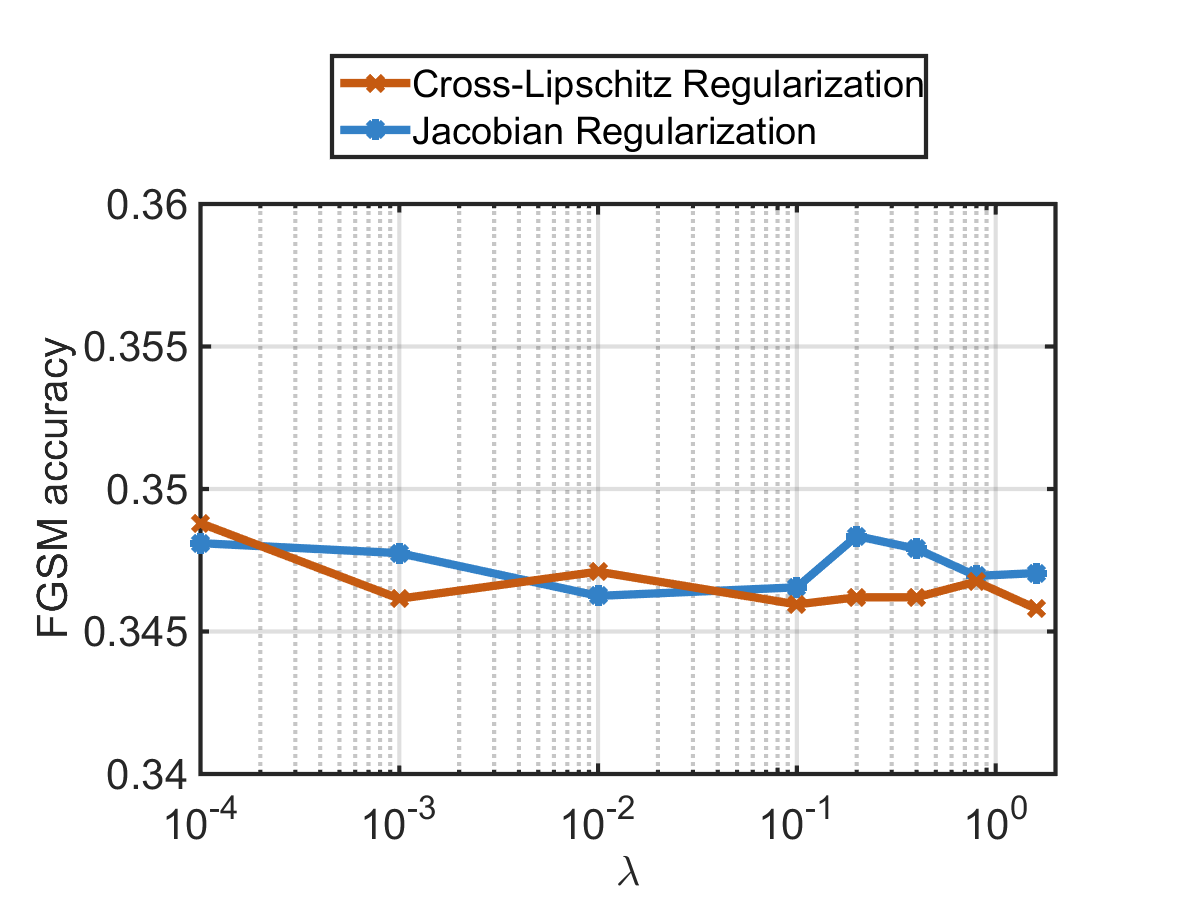}	
		\caption{}	
	\end{subfigure}
	\begin{subfigure}[b]{0.239\linewidth}
		\includegraphics[trim={0.0in 0.02in 0.5in 0.23in},clip, width=\linewidth]{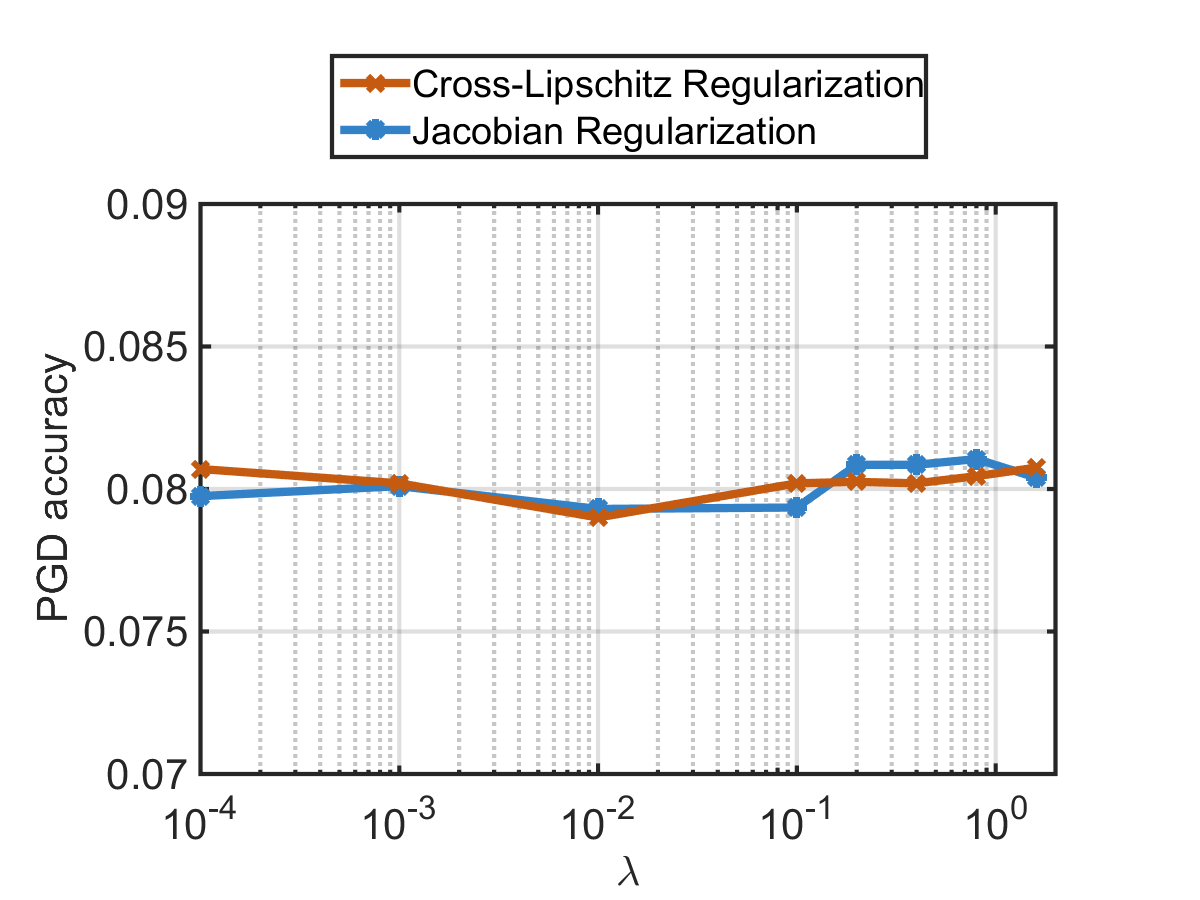}
		\caption{}
	\end{subfigure}
	\begin{subfigure}[b]{0.245\linewidth}
		\includegraphics[trim={0.0in 0.05in 0.5in 0.3in},clip, width=\linewidth]{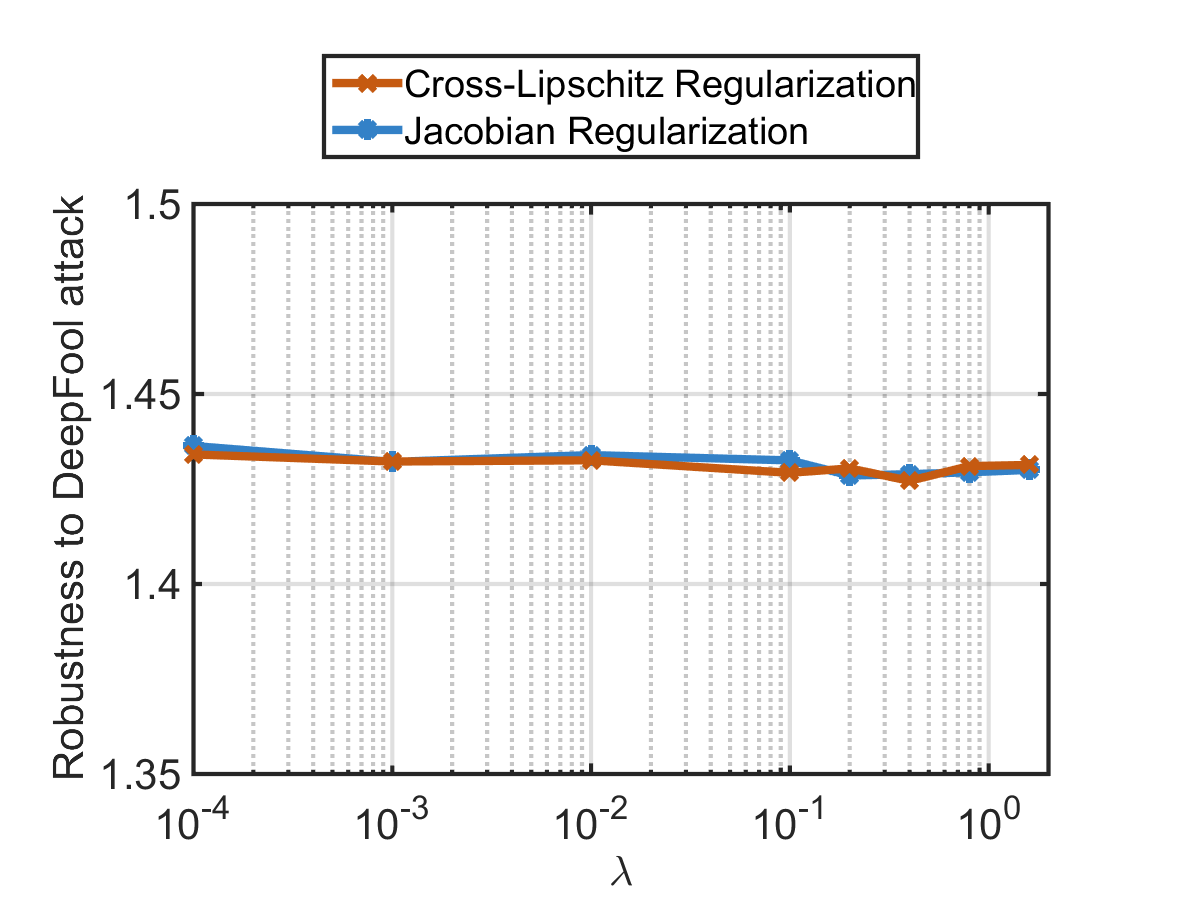}	
		\caption{}	
	\end{subfigure}
	\begin{subfigure}[b]{0.24\linewidth}
		\includegraphics[trim={0.0in 0.02in 0.5in 0.23in},clip, width=\linewidth]{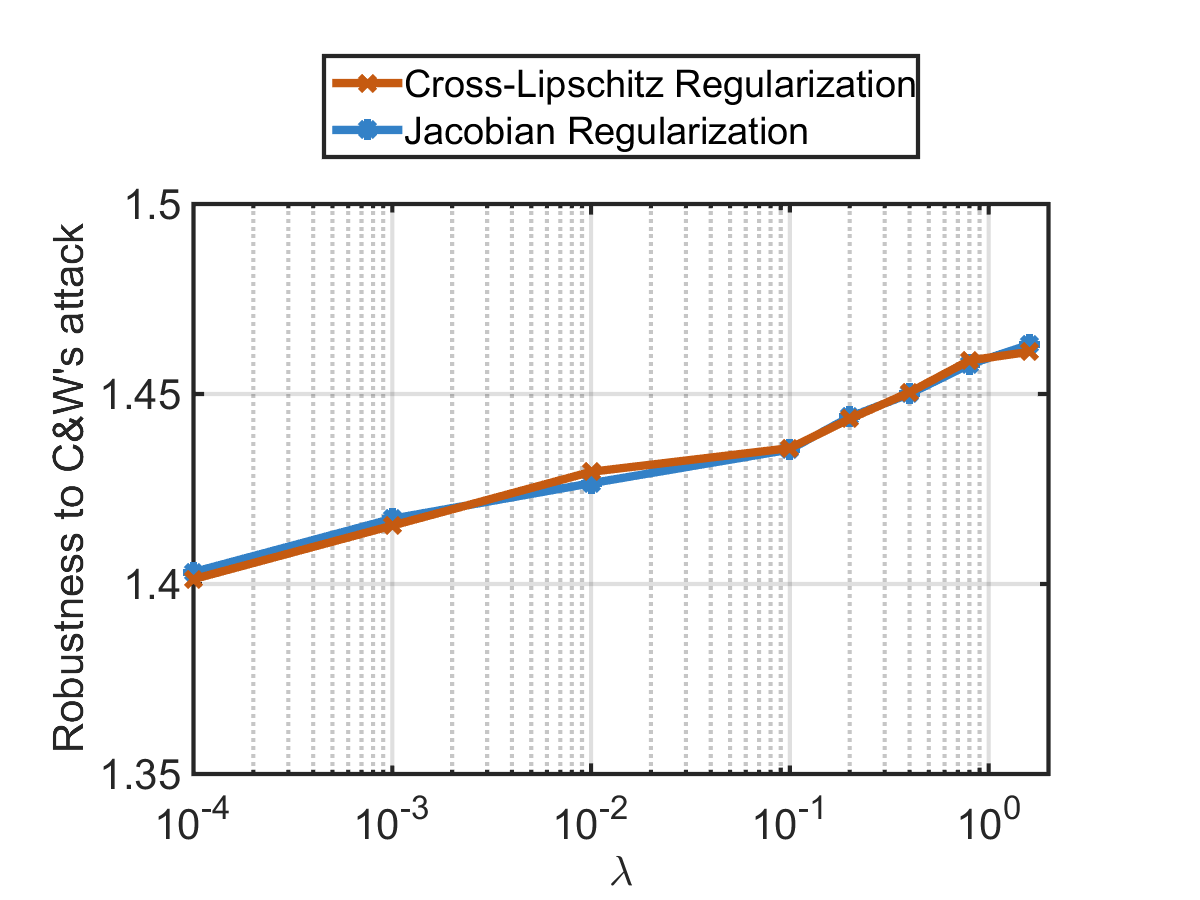}
		\caption{}
	\end{subfigure} \\

	\begin{subfigure}[b]{0.241\linewidth}
		\includegraphics[trim={0.0in 0.05in 0.5in 0.3in},clip, width=\linewidth]{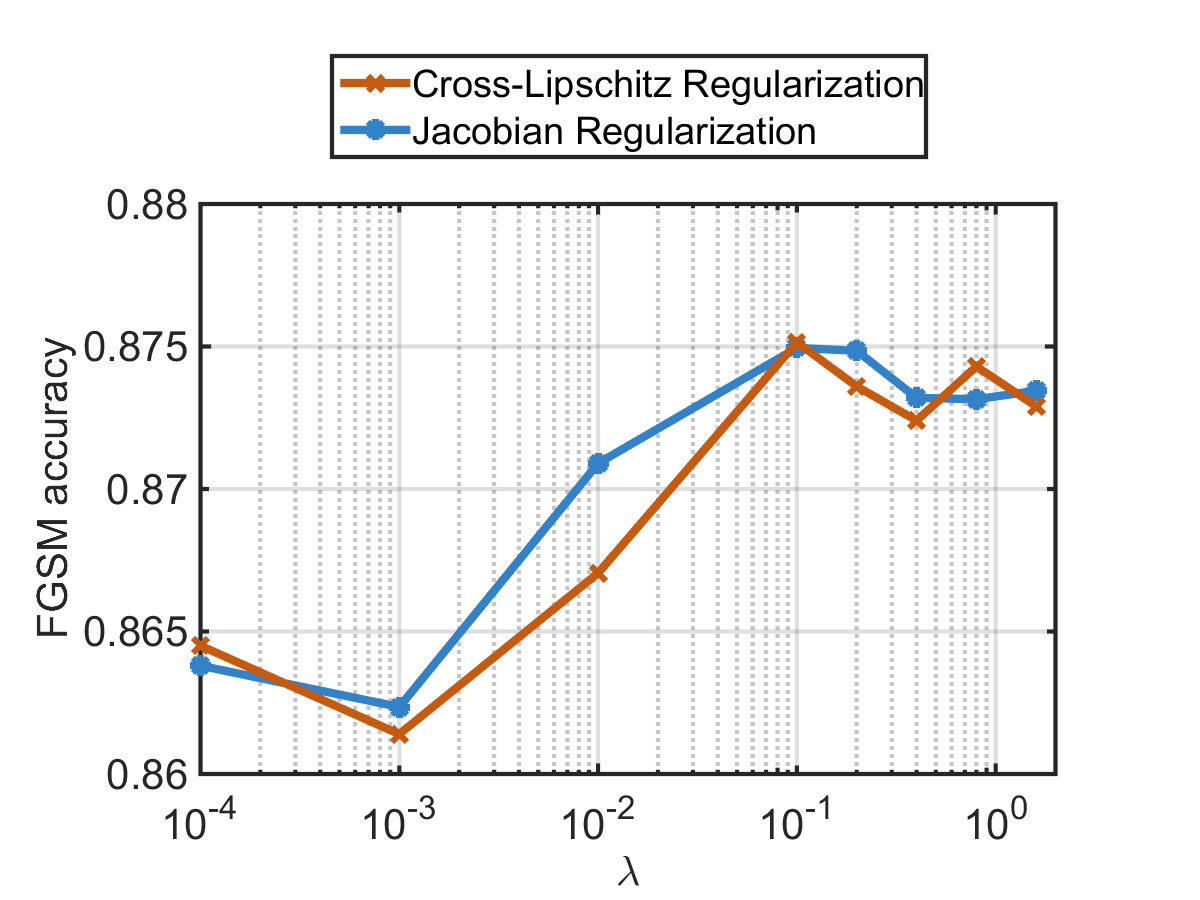}	
		\caption{}	
	\end{subfigure}
	\begin{subfigure}[b]{0.239\linewidth}
		\includegraphics[trim={0.0in 0.02in 0.5in 0.23in},clip, width=\linewidth]{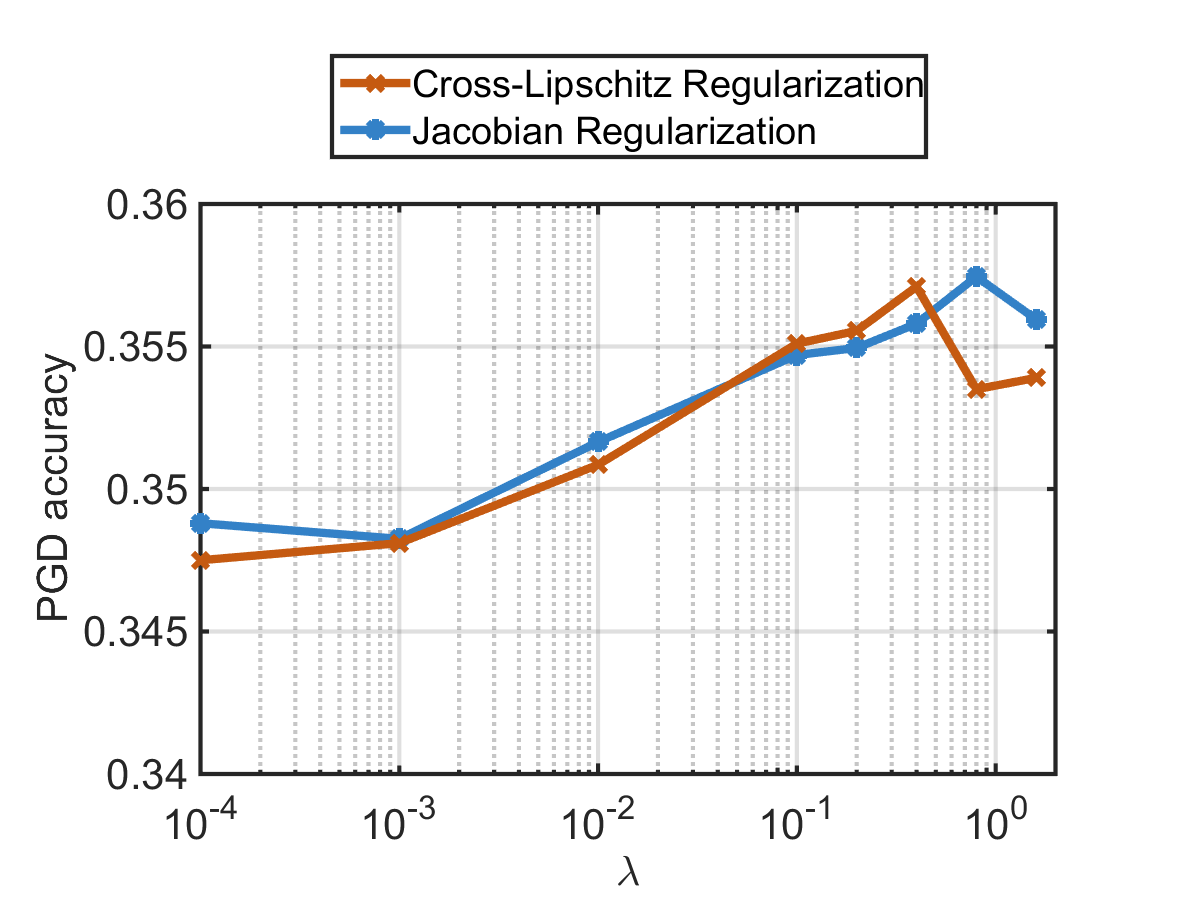}
		\caption{}
	\end{subfigure}
	\begin{subfigure}[b]{0.245\linewidth}
		\includegraphics[trim={0.0in 0.05in 0.5in 0.3in},clip, width=\linewidth]{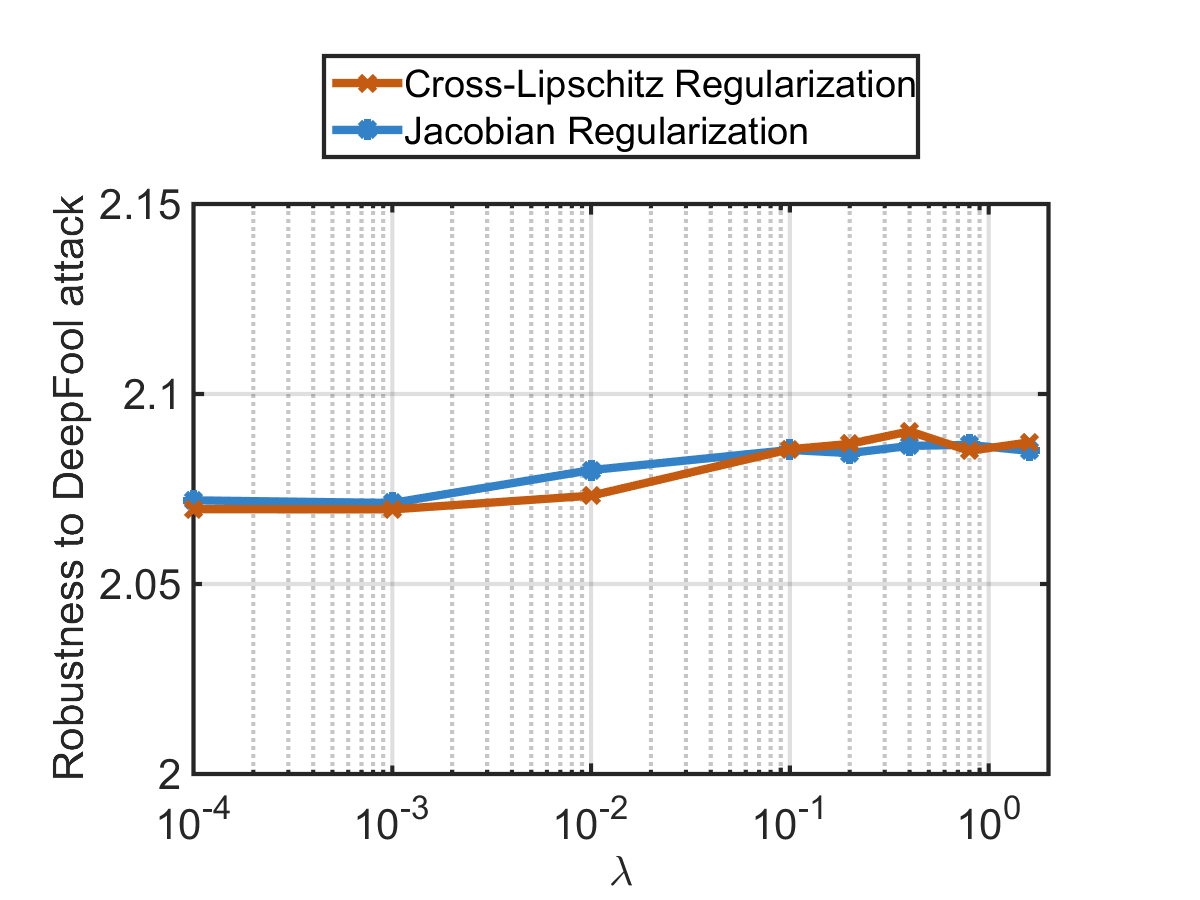}	
		\caption{}	
	\end{subfigure}
	\begin{subfigure}[b]{0.24\linewidth}
		\includegraphics[trim={0.0in 0.02in 0.5in 0.23in},clip, width=\linewidth]{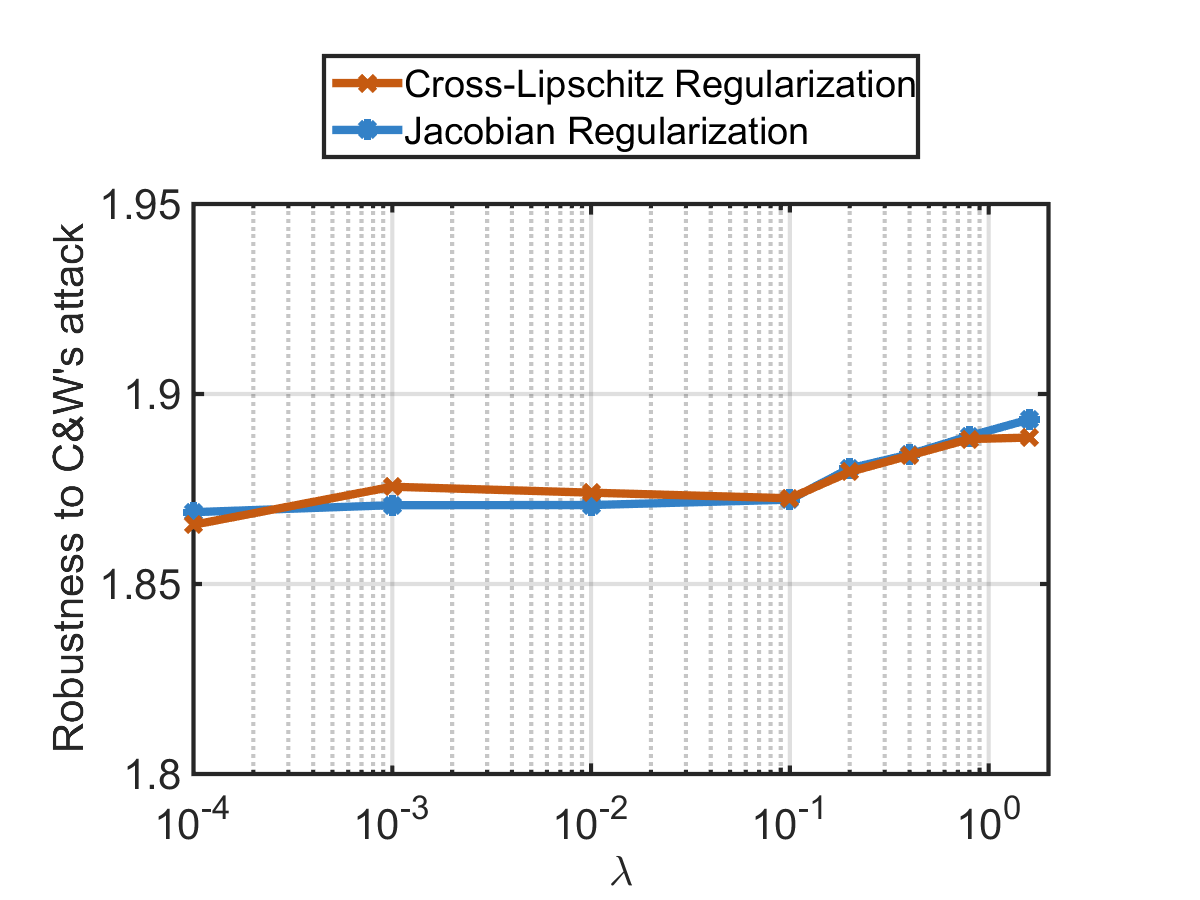}
		\caption{}
	\end{subfigure}\vskip -0.02in
	\caption{The adversarial robustness of obtained binary classification models evaluated with FGSM, PGD, DeepFool, and C\&W's attacks: (a)-(d) for LeNet-300-100 and (e)-(h) for LeNet-5. Ten runs from different initializations were performed and the average results are illustrated for fair comparisons. The y axis of the four subfigures on the left are normalized to the same numerical scale, and so as the four on the right. It can be seen that penalizing $\nu^2/2$ and $\mu^2$ perform similarly.}\vskip -0.1in  
\label{fig:3}
\end{figure*}

\begin{proposition}\textnormal{(An analytic expression for $\|\tilde{\mathbf r}^\ast\|_\infty$).}\label{pro:1.1}
For the binary classifier with a locally linear $g(\cdot)$ and a correctly classified instance $\mathbf x$, we have 
\begin{equation}
\|\tilde{\mathbf r}^*\|_\infty =  \frac{1}{p(\mathbf x)_y\|\mathbf v_{+}-\mathbf v_{-}\|_1}\left(\sqrt{1+\frac{2p(\mathbf x)_y\xi}{1-p(\mathbf x)_y}} - 1\right).
\end{equation}
\end{proposition}

\begin{proposition}\textnormal{(An upper bound of $\mathcal \eta^\ast$).}\label{pro:1.2}
For the binary classifier with a locally linear $g(\cdot)$ and a correctly classified input instance $\mathbf x$, we have $\forall \mathbf r\in \mathbb R^n$ satisfying $\|\mathbf r\|_\infty\leq \epsilon$, it holds that  \vskip -0.15in
\begin{equation}
\begin{aligned}
\mathcal \eta^\ast &= \mathcal L(\mathbf x, y) + \epsilon (1-p(\mathbf{x})_y)\|\mathbf v_{+} - \mathbf v_{-}\|_1 \\
&\quad\quad\quad+ \frac{1}{2}\epsilon^2 p(\mathbf{x})_y(1-p(\mathbf{x})_y)\|\mathbf v_{+} - \mathbf v_{-}\|_1^2.
\end{aligned}
\end{equation}
\end{proposition}

\subsection{Regularizations in Binary Classification}\label{sec:bcreg}
Besides Proposition~\ref{pro:1}, some intriguing corollaries can also be derived from Lemma~\ref{lem:2}. 
First, the direction of the input-gradient vector $\nabla$ is the same as that of the first eigenvector (\ie, the one corresponding to the largest eigenvalue) of matrix $H$. 
Second, we can derive $\|\nabla \|_2= (1-p(\mathbf x)_y)\nu$ and $\|H\|_F = \|H\|_2 = p(\mathbf x)_y(1-p(\mathbf x)_y)\nu^2$, which means that we further have simple analytic expressions for the concerned regularizers as:  
\begin{equation}\label{eq:9}
\begin{split}
\textbf{Jacobian regularizer} &:= \lambda\mu^2, \\
\textbf{Input-gradient regularizer}  &:= \lambda(1-p(\mathbf x)_y)^2\nu^2,  \\
\textbf{Curvature regularizer} &:= \lambda p(\mathbf x)_y(1-p(\mathbf x)_y) \nu^2, \\
\end{split}
\end{equation}
in which $\|H\|_2$ calculates the spectral norm (\ie, the matrix $l_2$ norm) of $H$, $\lambda>0$ is a hyper-parameter, and $\mu$ denotes $\|V\|_F$ which is apparently a local Lipschitz constant of $g(\cdot)$~\cite{Hein2017}. 
Third, it holds that $(1-p(\mathbf x)_y)^2\nu^2\leq p(\mathbf x)_y(1-p(\mathbf x)_y) \nu^2\leq\mu^2/2$ (\ie, $\|\nabla\|_2^2\leq\|H\|_2\leq\|V\|_F^2/2$), and thus we get a chained inequality of the regularizers.  
Without loss of generality, we write the regularizers in squared forms in Eq.~(\ref{eq:9}) for direct comparison.

One might have noticed that $\nu$ and $p(\mathbf x)_y$ are also the only ingredients in two of the regularizers in Eq.~\eqref{eq:9}.
In the remainder of this subsection, we shall discuss and highlight that: (1) the input-gradient regularization and curvature regularization both enforce suppression of $\nu^2$, which is in principle consistent with a cross-Lipschitz regularization~\cite{Hein2017}; (2) though the Jacobian regularization focuses on $\mu^2$ instead of $\nu^2$, there probably exists an underlying equivalence between penalizing scaled $\nu^2$ and $\mu^2$; (3) critical discrepancies still exist amongst these regularizations, mostly about $p(\mathbf x)_y$.

\textbf{Cross-Lipschitz vs. Lipschitz:} With clear expressions in Eq.~(\ref{eq:7}) and~(\ref{eq:9}), we know that the input-gradient regularization and curvature regularization are similar to a \emph{cross-Lipschitz regularization} that penalizes $\lambda\nu^2/2$~\cite{Hein2017}~\footnote{Interested readers can refer to Section G for rigorous analyses.}, while the Jacobian regularization penalizes $\lambda\mu^2$ (with a local \emph{Lipschitz} constant $\mu$) and it boils down to weight decay in single-layer perceptrons and linear classifiers.
Although it seems as if the Jacobian regularization was different from the others, in light of the Parseval tight frame and Parseval networks~\cite{Cisse2017}, we conjecture nonetheless that there exists an equivalence between penalizing scaled $\nu^2$ (as with the cross-Lipschitz regularization, input-gradient regularization, and curvature regularization) and $\mu^2$ (as with the Jacobian regularization). 
To shed light on this, more discussions are performed as follows. 

First and foremost, it is self-evident that the inequality  
\begin{equation}
\nu^2/2 \leq \frac{1}{2}\|\mathbf v_{+}\|^2_2 + |\mathbf v^T_{+}\mathbf v_{-}| + \frac{1}{2}\|\mathbf v_{-}\|^2_2 \leq \mu^2
\end{equation}
holds, thus one might argue that adopting the Jacobian regularization also implies small $\nu^2$ in obtained models as with the cross-Lipschitz regularization.
Second, for single-layer perceptrons, we can easily verify that the function $g(\cdot)$ is convex, and thus the Jacobian regularized training loss is strongly convex w.r.t. $V$. 
Considering that the columns of $V$ can be processed simultaneously by adding/subtracting a vector whilst the classification decision and cross-entropy loss won't change, we have $-\mathbf v_{+}=\mathbf v_{-}$ for the optimal $V$ and an equivalence is achieved between penalizing $\lambda\nu^2/2$ and $\lambda\mu^2$ through derivation.
The result naturally generalizes to DNNs with locally linear $g(\cdot)$ (\ie, DNNs with general ReLU activations) of our interest, if only the final layer is to be optimized. 
The following proposition makes this formal, and the proof can be found in Appendix D. 
\begin{proposition}\textnormal{(A derived equivalence).}\label{pro:2}
For a single-layer perceptron or a piecewise linear DNN in which only the final layer parameterized by $W_d$ is to be optimized, we have the equivalence:
\begin{equation}\label{eq:11}
\begin{aligned}
\forall \lambda\geq0, \quad &\argmin_{W_d} \mathbb E_{(\mathbf x, y)}[\mathcal L(\mathbf x, y; V)+\lambda \nu^2/2] = \\
&\quad\quad\quad\argmin_{W_d} \mathbb E_{(\mathbf x, y)}[\mathcal L(\mathbf x, y; V)+\lambda \mu^2].
\end{aligned}
\end{equation}
\end{proposition}
\vskip -0.05in
In addition to the above results, we further show that the two regularizations can lead to the same gradient flow in certain scenarios. 
One example in which this can be demonstrated is when the first feature is uncorrelated with the label $y$ and the other $(n-1)$ features are distributed normally with the mean value being propotional to $y$ (\ie, they are weakly correlated with the label)~\cite{Tsipras2019}. 
We let $\mathbf v_{+}\leftarrow[0, a,\ldots, a]$ and $\mathbf v_{-}\leftarrow[0, -a,\ldots, -a]$ approach the Bayes error rate.
Under such circumstance, the two regularizations initialized from the Bayes classifier share the same gradient flow for their $V$ matrices, provided $2\times$ smaller penalty to $\nu^2$ than to $\mu^2$ as in Eq.~(\ref{eq:11}) .

To test whether the revealed equivalence generalizes to practical scenarios, we conducted an experiment on distinguishing the digit ``7'' from ``1'' using MNIST images.
Our experimental settings and many more details are carefully introduced in Appendix F. 
As suggested~\cite{Jakubovitz2018,Moosavi2019}, we first trained baseline models from scratch without any explicit regularization, then fine-tuned the models using different regularization strategies and evaluated the obtained adversarial robustness to FGSM~\cite{Goodfellow2015}, PGD~\cite{Madry2018}, DeepFool~\cite{Moosavi2016}, and the C\&W's attack~\cite{Carlini2017}. 
We trained MLPs and convolutional networks with ReLU nonlinearity following the “LeNet-300-100” and “LeNet-5” architectures in prior work~\cite{Lecun1998}.
Figure~\ref{fig:3} compares the performance of regularizations incorporating $\lambda\nu^2/2$ and $\lambda\mu^2$. 
With varying $\lambda$, it can be seen that the regularized models show similar robustness in almost all test cases.
Similar results on CIFAR-10 with ResNets and VGG-like networks can be found in Appendix F. 

\begin{figure}[h!]
	\centering
	\vspace{-0.1in}
	\includegraphics[width=0.6\linewidth, trim=0.2in 0 0.2in 0.1in, clip]{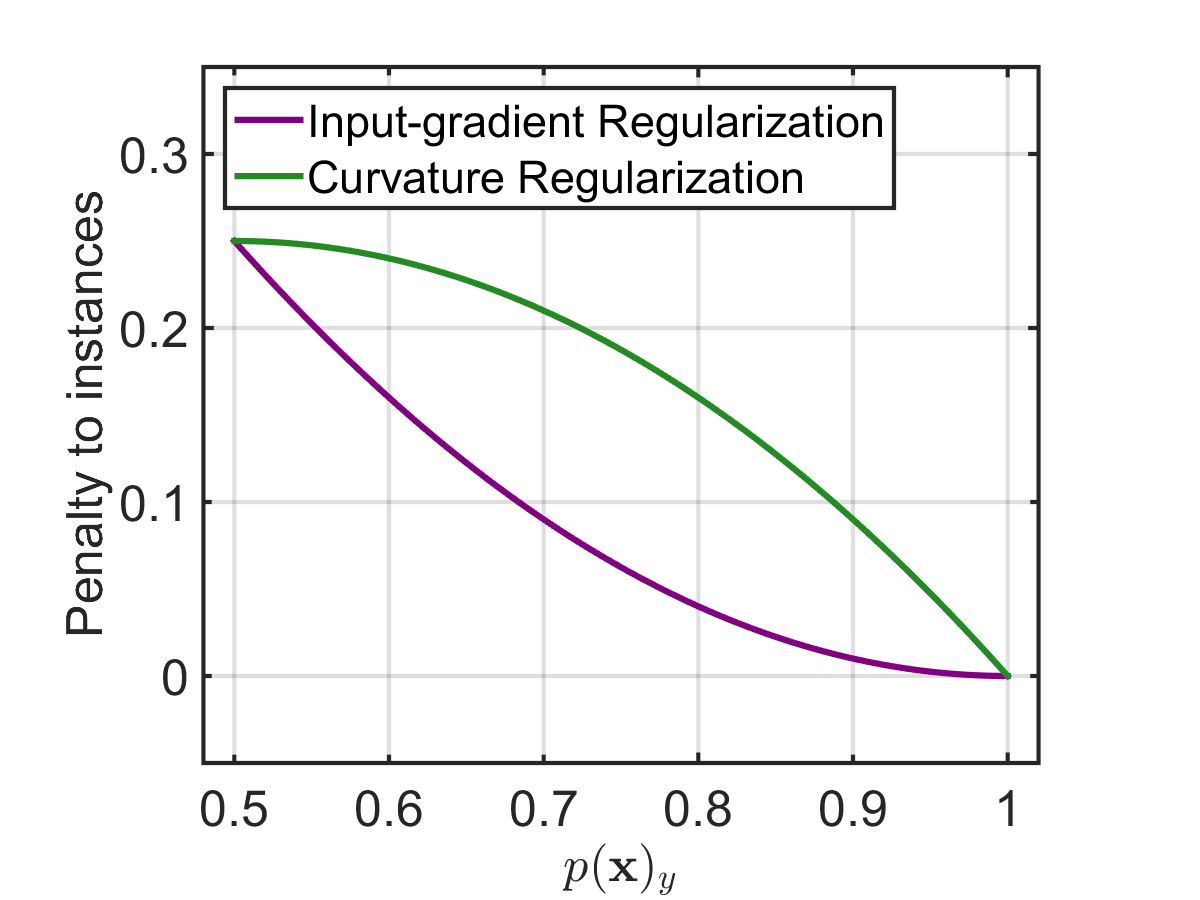}
	\vskip -0.5em
	\caption{Different regularizers focus on samples with different prediction confidence.}
	\label{fig:2}
\end{figure}

\begin{figure*}[t]
	\centering 
	\begin{subfigure}[b]{0.245\linewidth}
		\includegraphics[trim={0.0in 0.25in 0.5in 0.3in},clip, width=\linewidth]{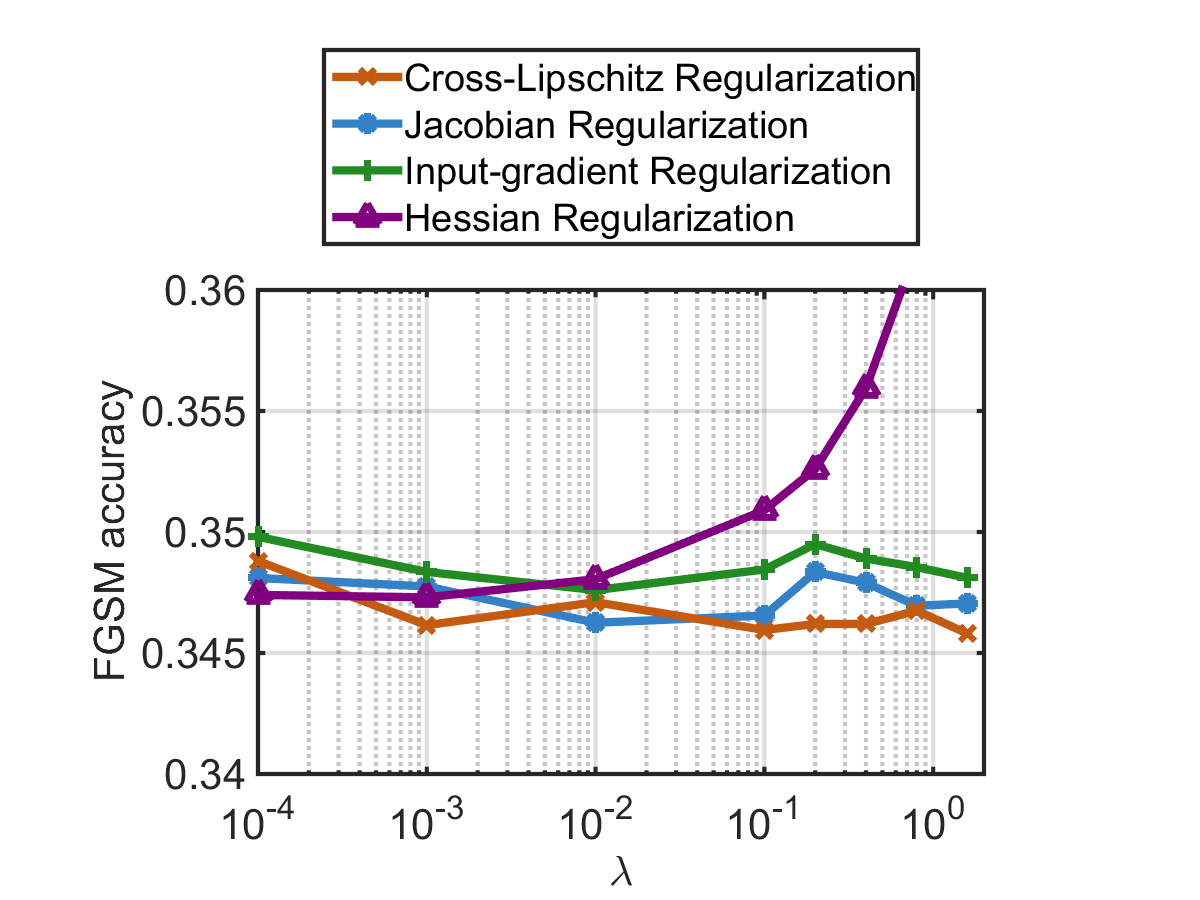}	
		\caption{}	
	\end{subfigure}
	\begin{subfigure}[b]{0.243\linewidth}
		\includegraphics[trim={0.0in 0.2in 0.5in 0.23in},clip, width=\linewidth]{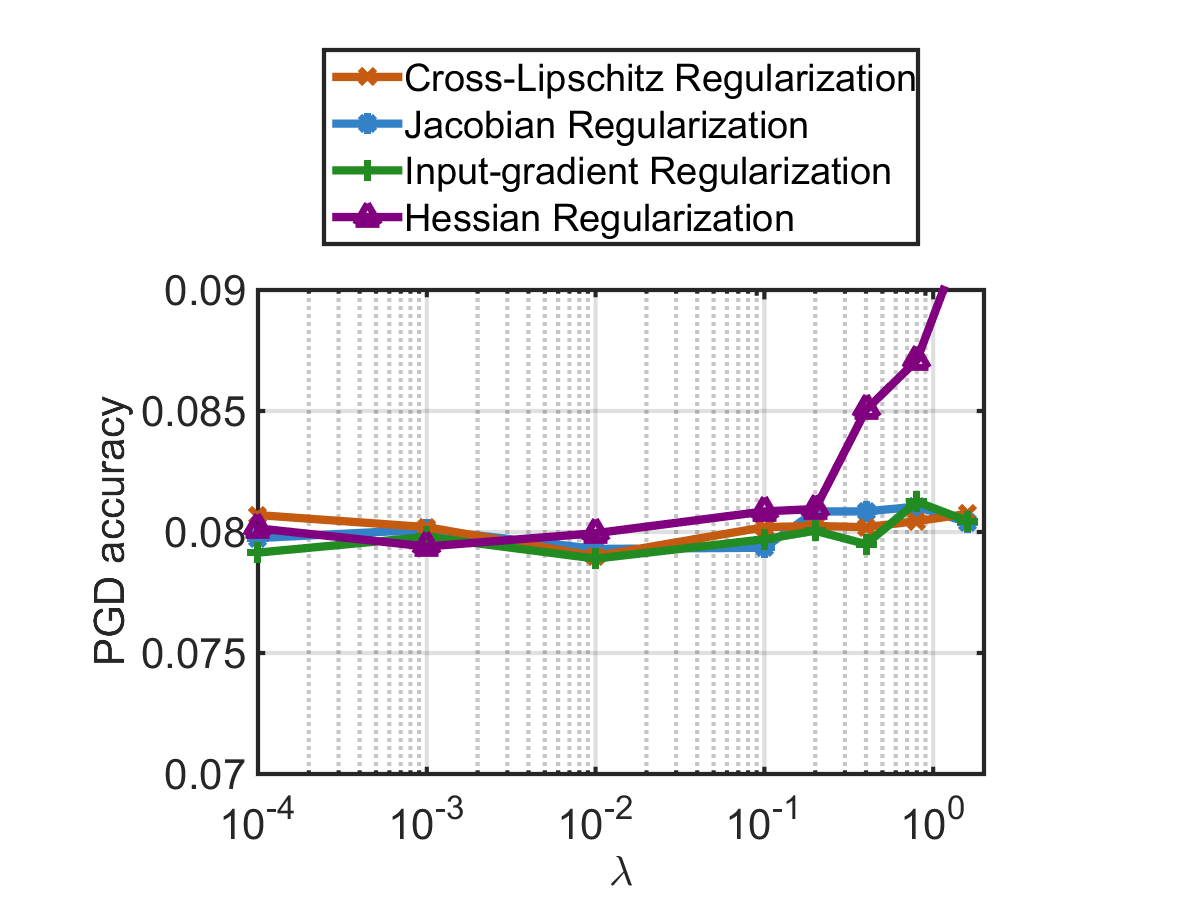}
		\caption{}
	\end{subfigure}
	\begin{subfigure}[b]{0.249\linewidth}
		\includegraphics[trim={0.0in 0.05in 0.5in 0.3in},clip, width=\linewidth]{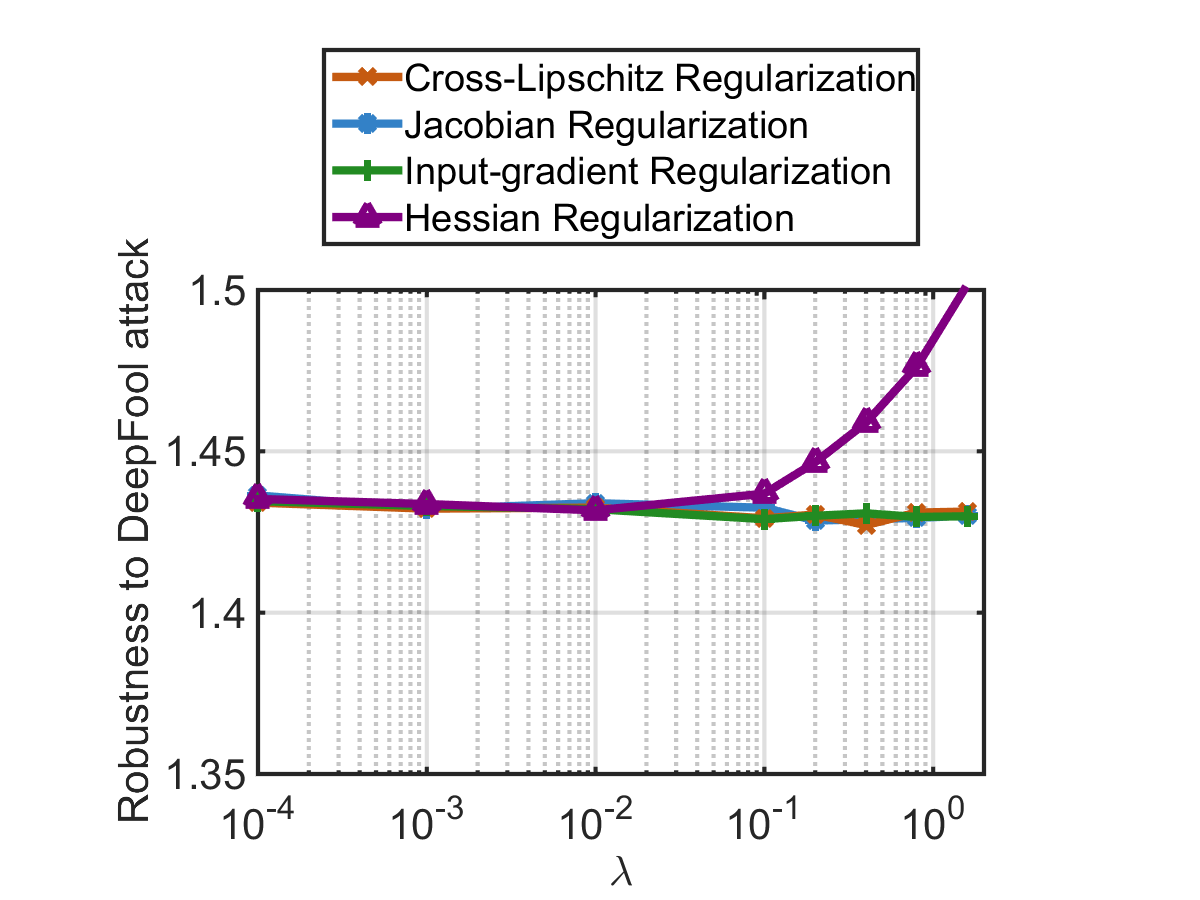}	
		\caption{}	
	\end{subfigure}
	\begin{subfigure}[b]{0.244\linewidth}
		\includegraphics[trim={0.0in 0.02in 0.5in 0.23in},clip, width=\linewidth]{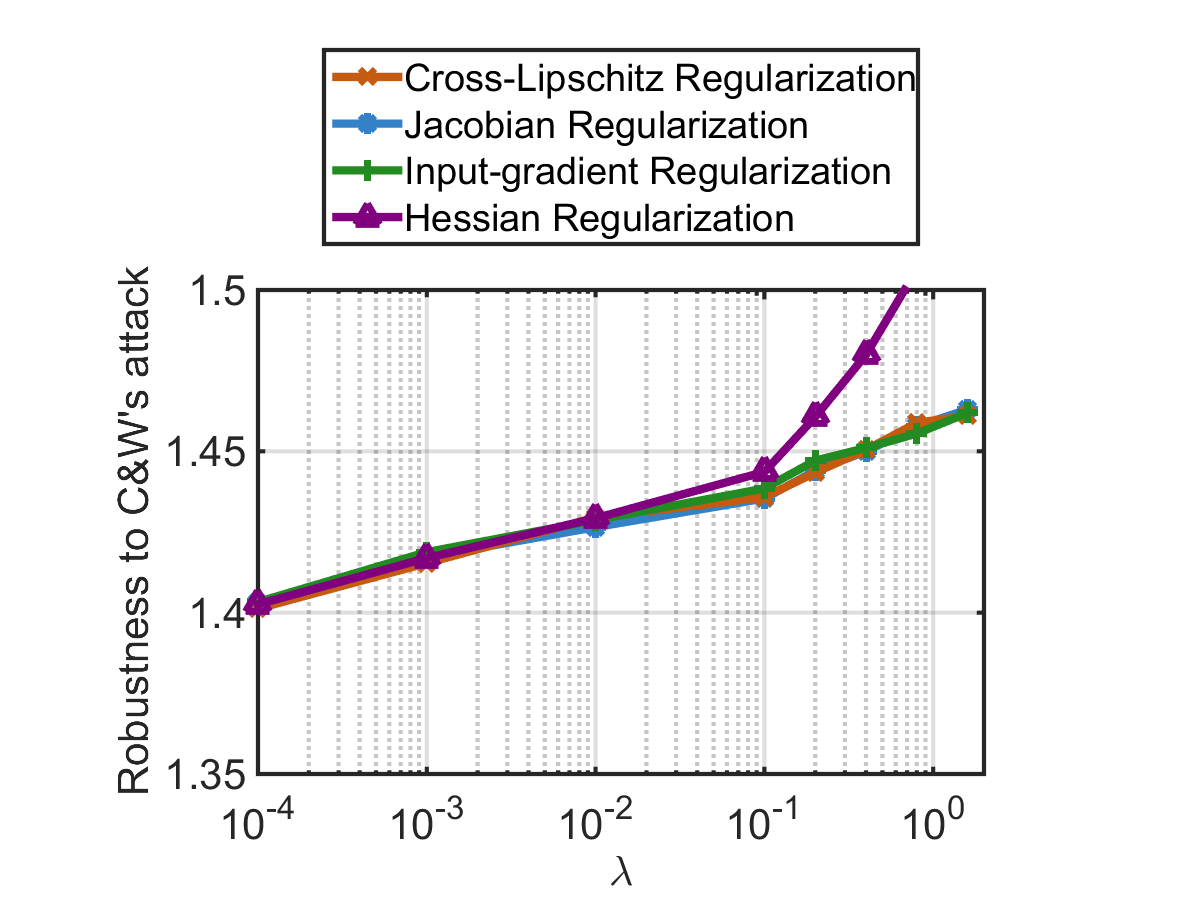}
		\caption{}
	\end{subfigure} \\

	\begin{subfigure}[b]{0.245\linewidth}
		\includegraphics[trim={0.0in 0.05in 0.5in 0.3in},clip, width=\linewidth]{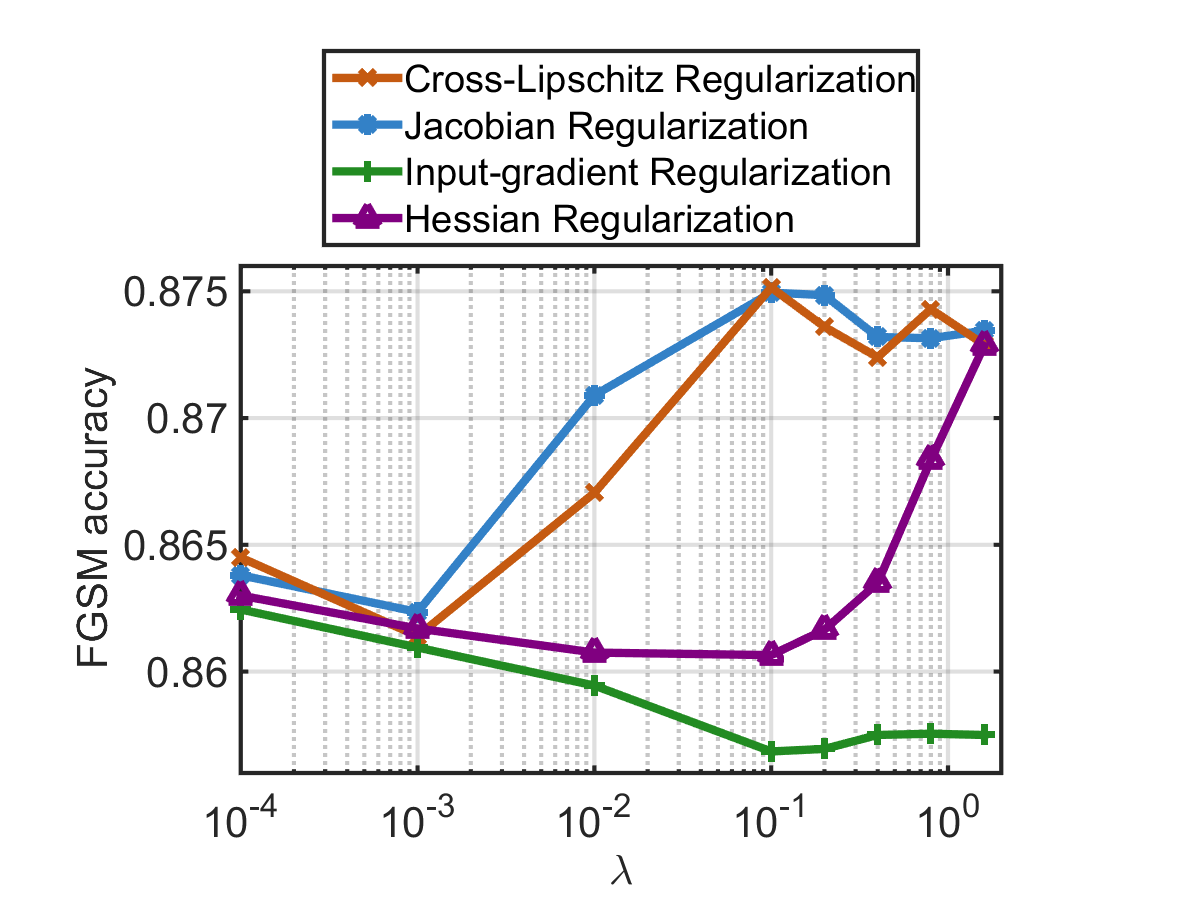}	
		\caption{}	
	\end{subfigure}
	\begin{subfigure}[b]{0.243\linewidth}
		\includegraphics[trim={0.0in 0.02in 0.5in 0.23in},clip, width=\linewidth]{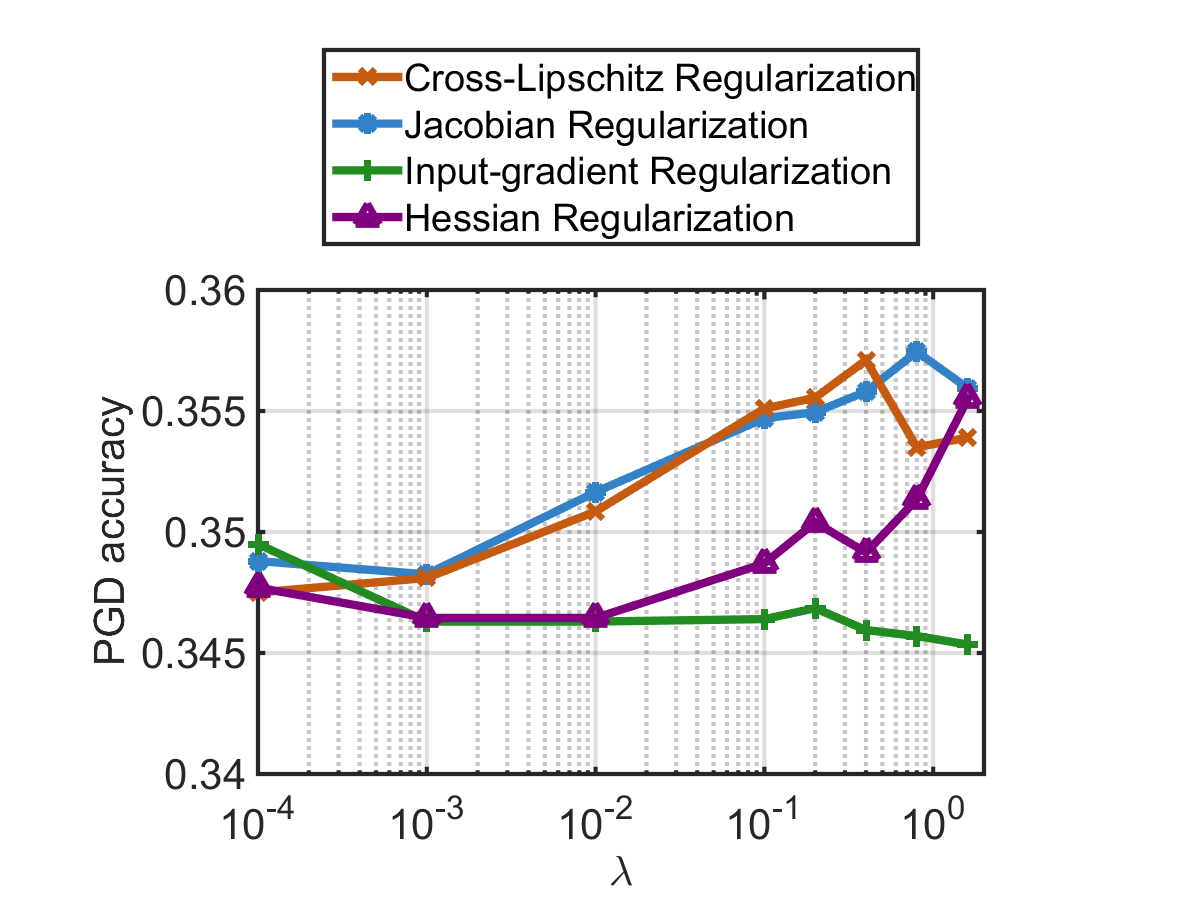}
		\caption{}
	\end{subfigure}
	\begin{subfigure}[b]{0.249\linewidth}
		\includegraphics[trim={0.0in 0.05in 0.5in 0.3in},clip, width=\linewidth]{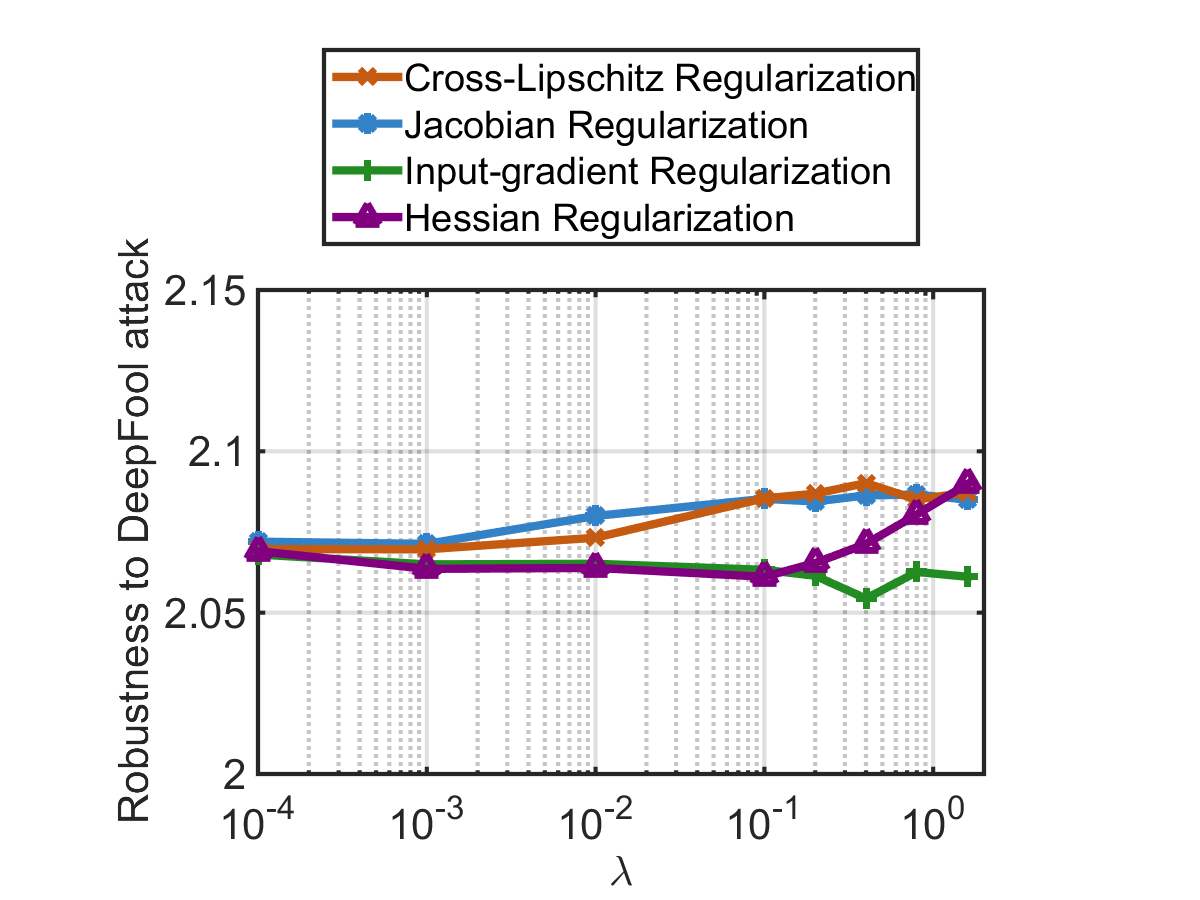}	
		\caption{}	
	\end{subfigure}
	\begin{subfigure}[b]{0.244\linewidth}
		\includegraphics[trim={0.0in 0.02in 0.5in 0.23in},clip, width=\linewidth]{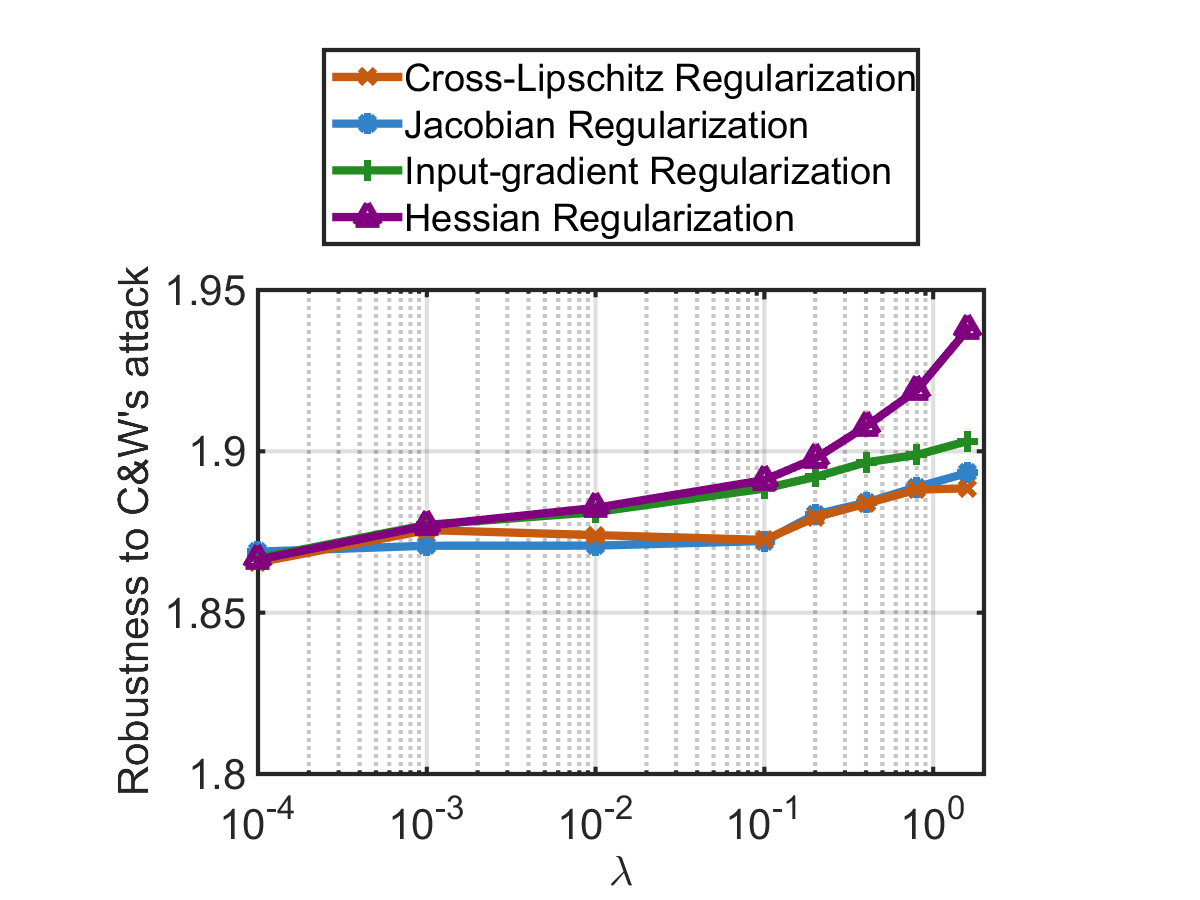}
		\caption{}
	\end{subfigure} \vskip -0.1in
	\caption{The robustness of \emph{all} obtained binary classification models evaluated with FGSM, PGD, DeepFool, and the C\&W's attacks: (a)-(d) for LeNet-300-100 and (e)-(h) for LeNet-5. Ten runs from different initializations were performed and the average results are reported. The y axis of the four subfigures on the left are normalized to the same numerical scale, and so as the four on the right. It can be seen that the curvature regularization shows the most promising performance.}\vskip -0.1in
\label{fig:4}
\end{figure*}

\textbf{``Confidence'' in regularizations:} Apart from suppressing the local (cross-)Lipschitz constants, the input-gradient regularizer and curvature regularizer both involve the prediction probability $p(\mathbf x)_y$ in Eq.~\eqref{eq:9}, with different objectives though.
By incorporating $(1-p(\mathbf x)_y)(1-p(\mathbf x)_y)$, the input-gradient regularization encourages model
predictions with high confidence. 
If $\nu^2$ is fixed, then the $p(\mathbf x)_y$-related term in the input-gradient regularizer acts as an additional prediction loss during training.
It has larger penalties and slopes (in absolute value) for the training instances with relatively smaller $p(\mathbf x)_y$, \ie, lower confidence.
Similarly, we know that the curvature regularization involves $p(\mathbf x)_y(1-p(\mathbf x)_y)$ and advocates large $p(\mathbf x)_y$ as well.
However, as depicted in the green curve in Figure~\ref{fig:2}, the function exhibits larger absolute value of slope at predictions with higher confidence, which is different from $p(\mathbf x)_y(1-p(\mathbf x)_y)$ but consistent with the preference of $\|\mathbf r^\ast\|_2$ as shown in Figure~\ref{fig:1} right.
As for the cross-Lipschitz regularizer and Jacobian regularizer, no $p(\mathbf x)_y$-related term is explicitly involved whatsoever.~\footnote{See Eq.~(\ref{eq:9}), the ``regularizer'' means the regularization term itself in this paper. Note that the cross-entropy term involves the prediction probability $p(\mathbf x)_y$ of course.} 

Although it is unclear which of the tactics would be the most suitable one in practice, one might be aware that different choices perform dis-similarly, otherwise we should have obtained functional equivalence for all these contestants. 
In order to figure out the best one in practice, we compared the achieved robustness via input-gradient regularization and curvature regularization empirically with our results using the cross-Lipschitz regularization and Jacobian regularization.
As shown in Figure~\ref{fig:4}, the lately developed curvature regularization surpasses all its competitors with reasonably large $\lambda$ values, showing the superiority of its specific tactic of handling confident predictions. 
Notice that we retain the same numerical ranges of axes in Figure~\ref{fig:4} as in Figure~\ref{fig:3}, but some newly drawn curves (for curvature regularization) in Figure~\ref{fig:4} may be too promising to stick in the plot.

\begin{figure*}[t]
	\centering 
	\begin{subfigure}[b]{0.245\linewidth}
		\includegraphics[trim={0.0in 0.05in 0.5in 0.3in},clip, width=\linewidth]{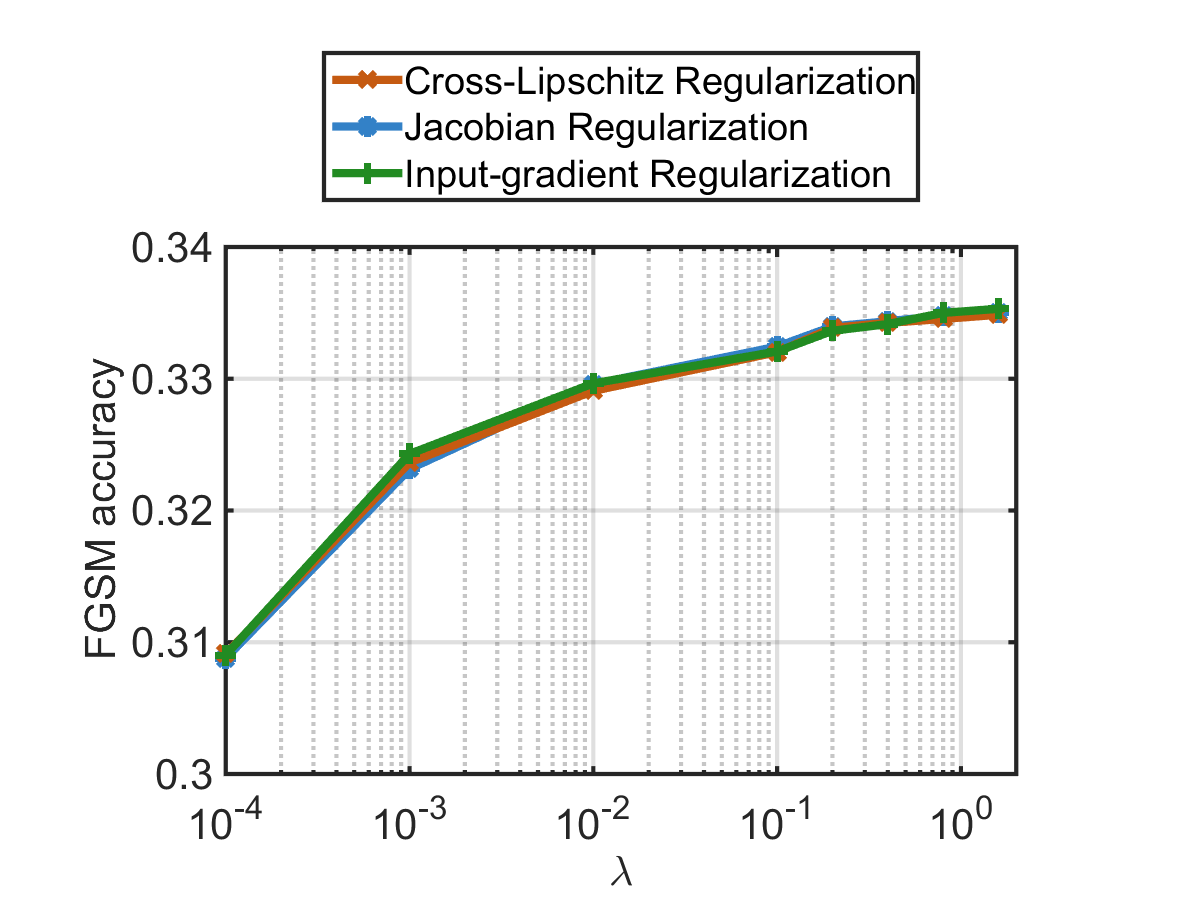}	
		\caption{}	
	\end{subfigure}
	\begin{subfigure}[b]{0.243\linewidth}
		\includegraphics[trim={0.0in 0.02in 0.5in 0.23in},clip, width=\linewidth]{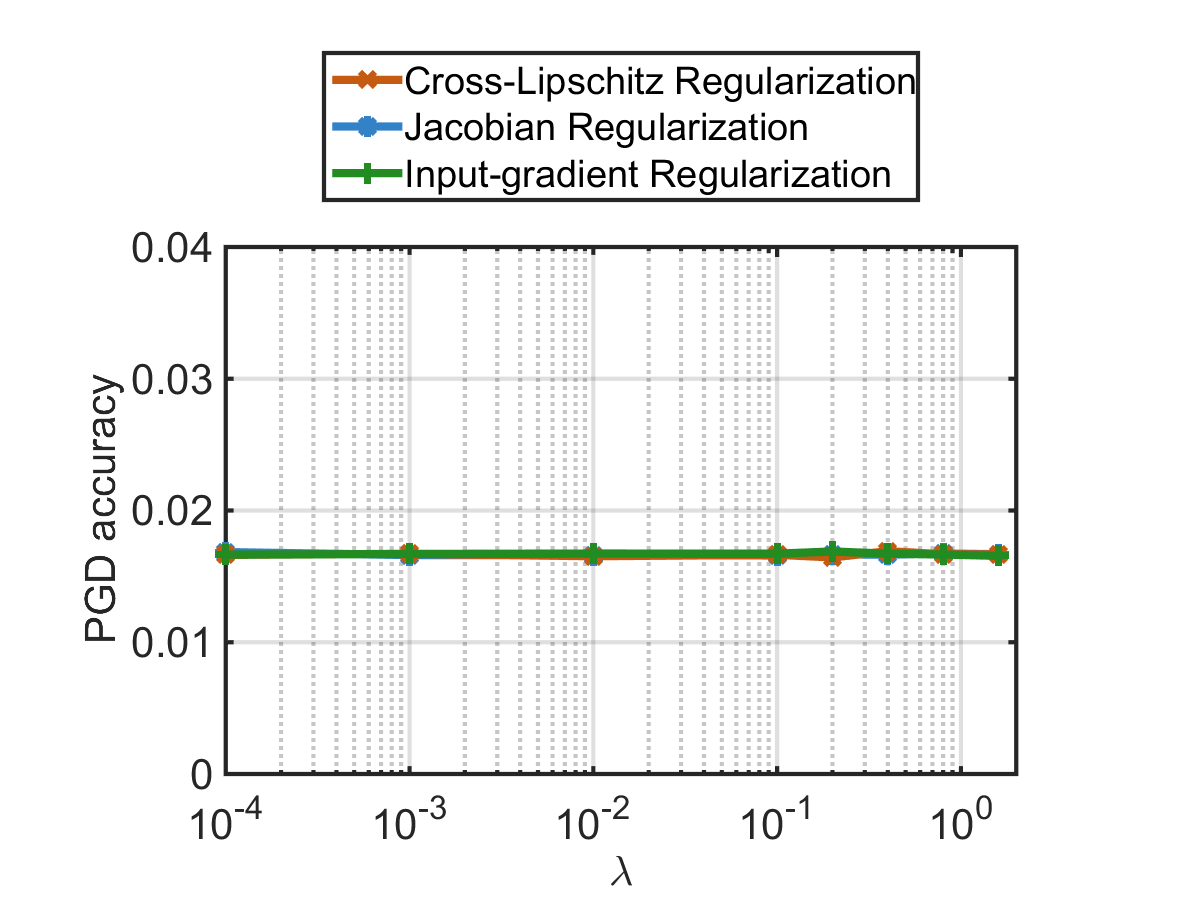}
		\caption{}
	\end{subfigure}
	\begin{subfigure}[b]{0.249\linewidth}
		\includegraphics[trim={0.0in 0.05in 0.5in 0.3in},clip, width=\linewidth]{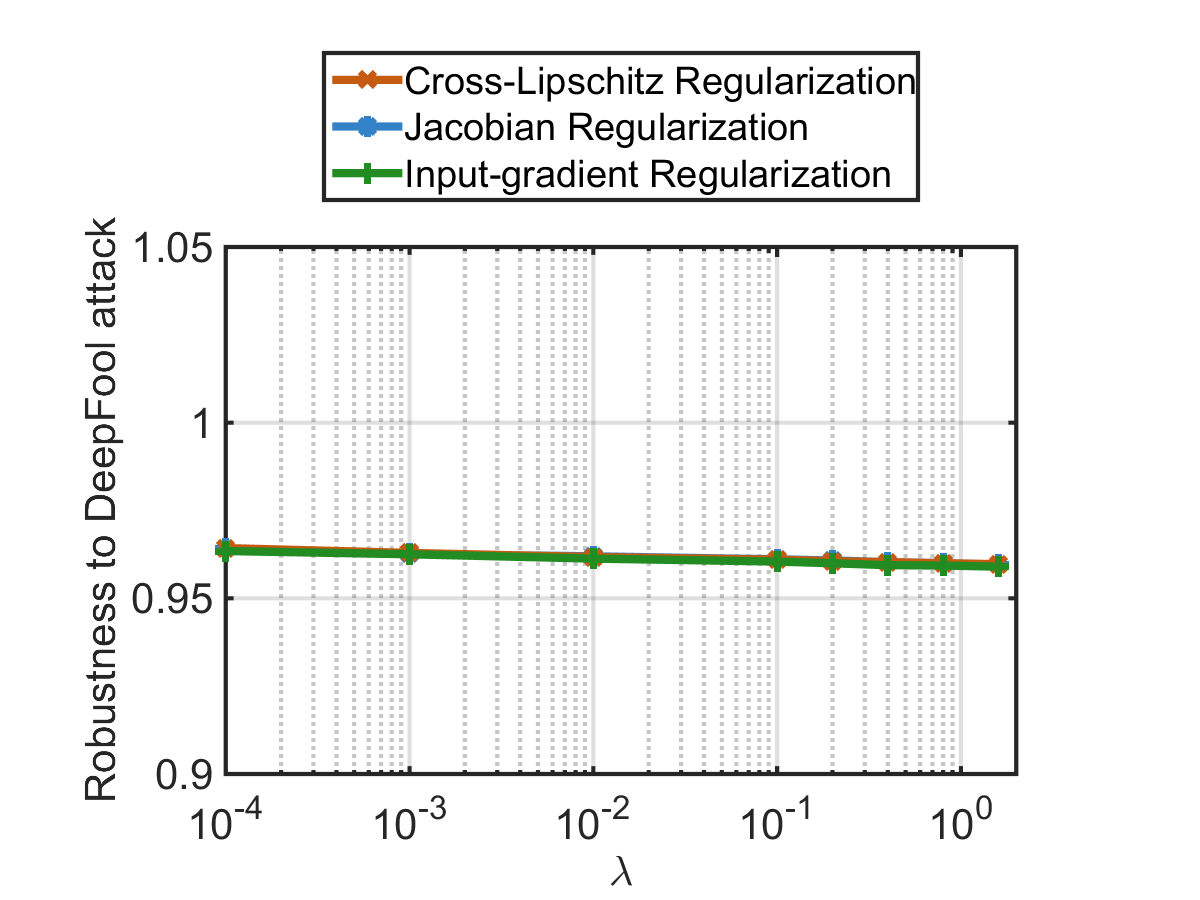}	
		\caption{}	
	\end{subfigure}
	\begin{subfigure}[b]{0.244\linewidth}
		\includegraphics[trim={0.0in 0.02in 0.5in 0.23in},clip, width=\linewidth]{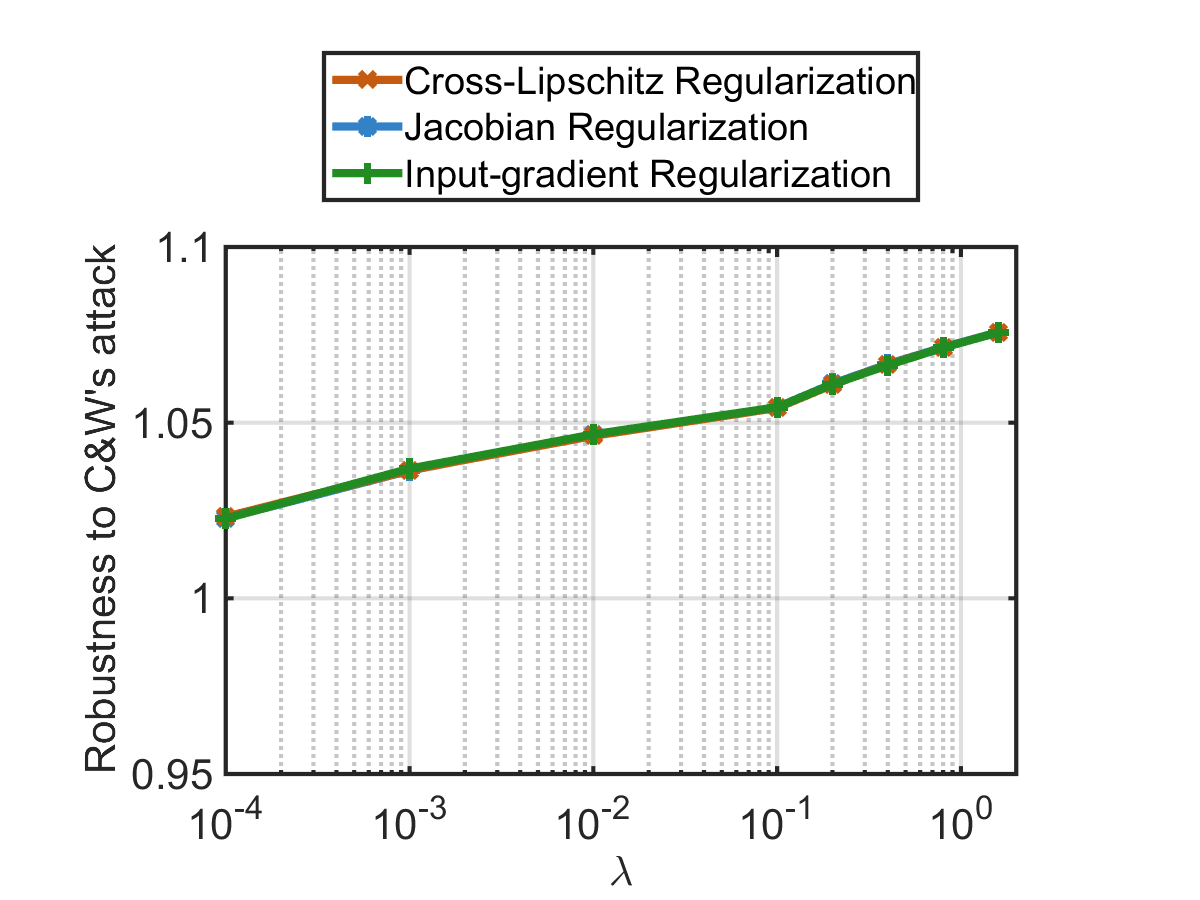}
		\caption{}
	\end{subfigure} \\

	\begin{subfigure}[b]{0.245\linewidth}
		\includegraphics[trim={0.0in 0.05in 0.5in 0.3in},clip, width=\linewidth]{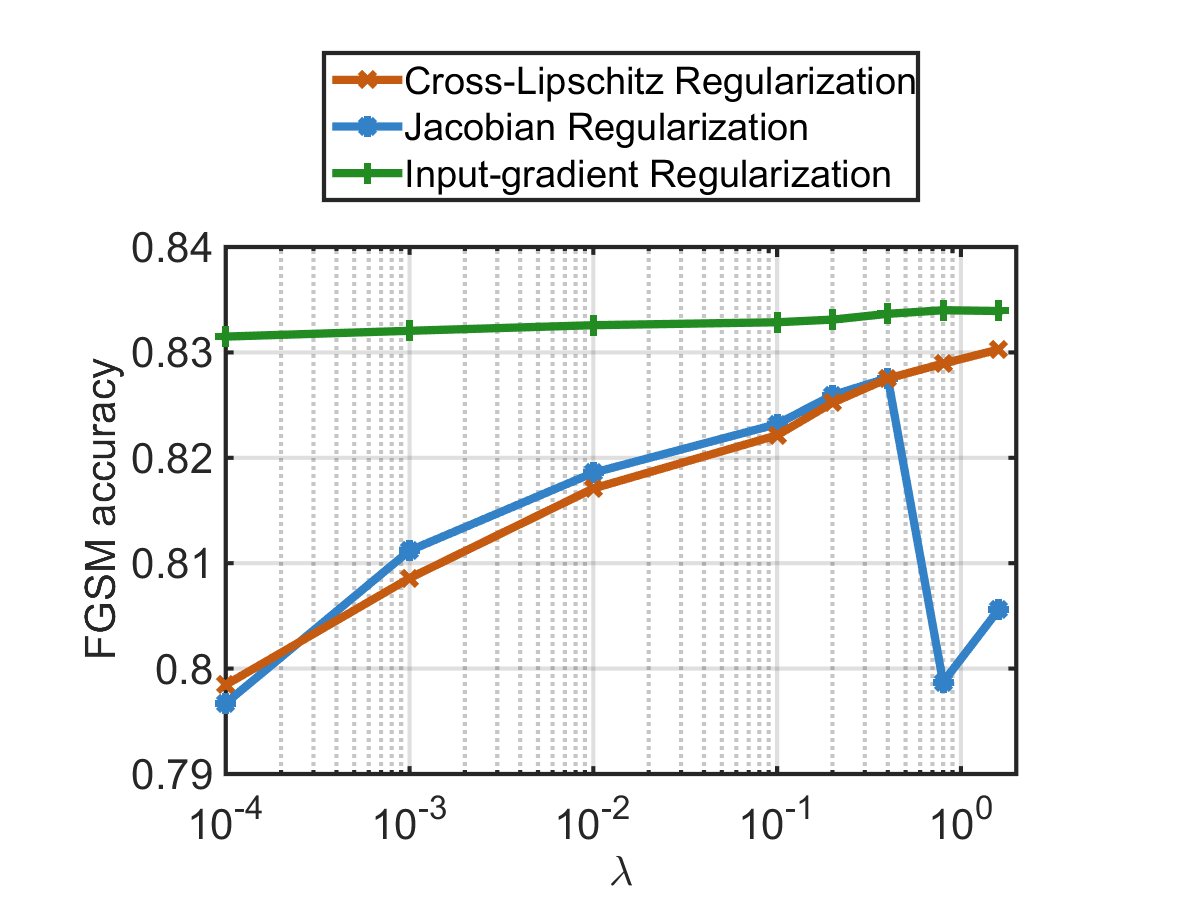}	
		\caption{}	
	\end{subfigure}
	\begin{subfigure}[b]{0.243\linewidth}
		\includegraphics[trim={0.0in 0.02in 0.5in 0.23in},clip, width=\linewidth]{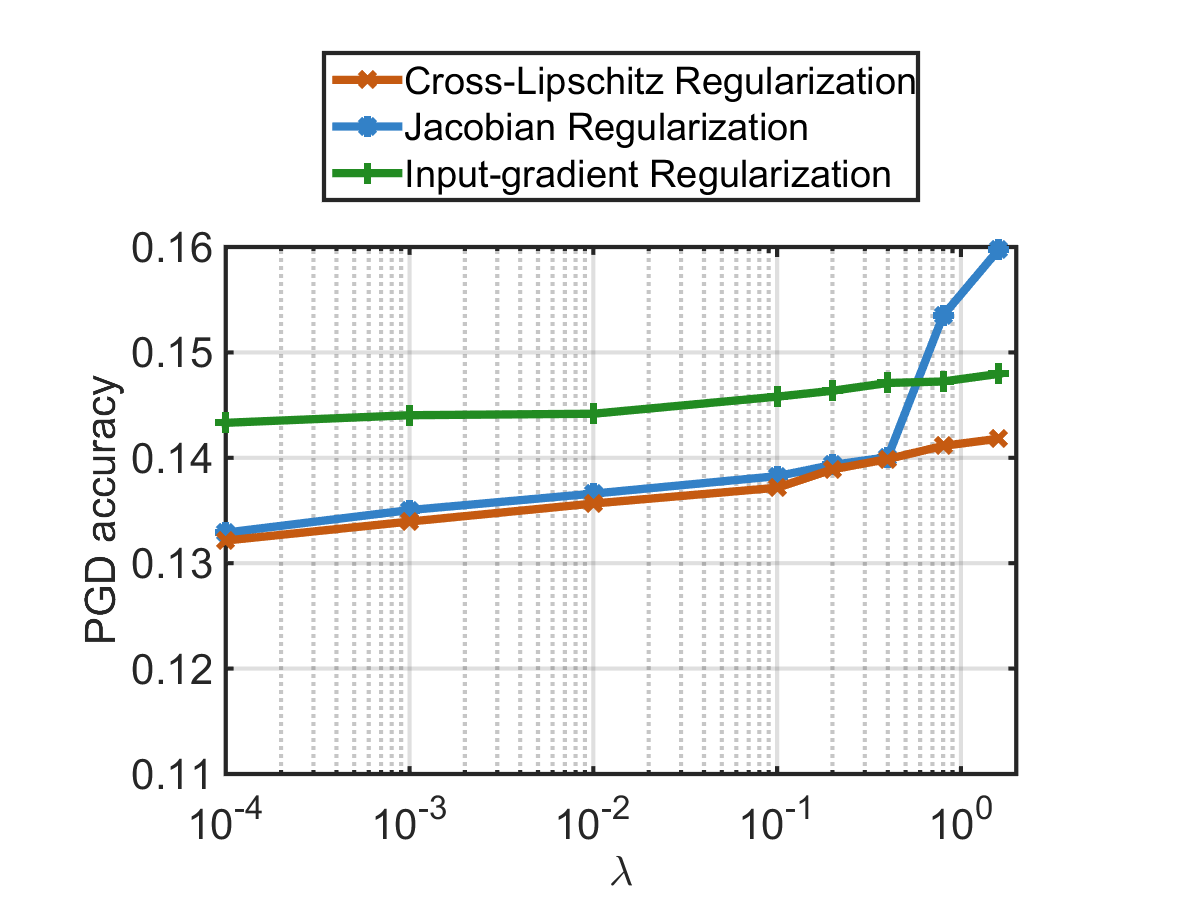}
		\caption{}
	\end{subfigure}
	\begin{subfigure}[b]{0.249\linewidth}
		\includegraphics[trim={0.0in 0.05in 0.5in 0.3in},clip, width=\linewidth]{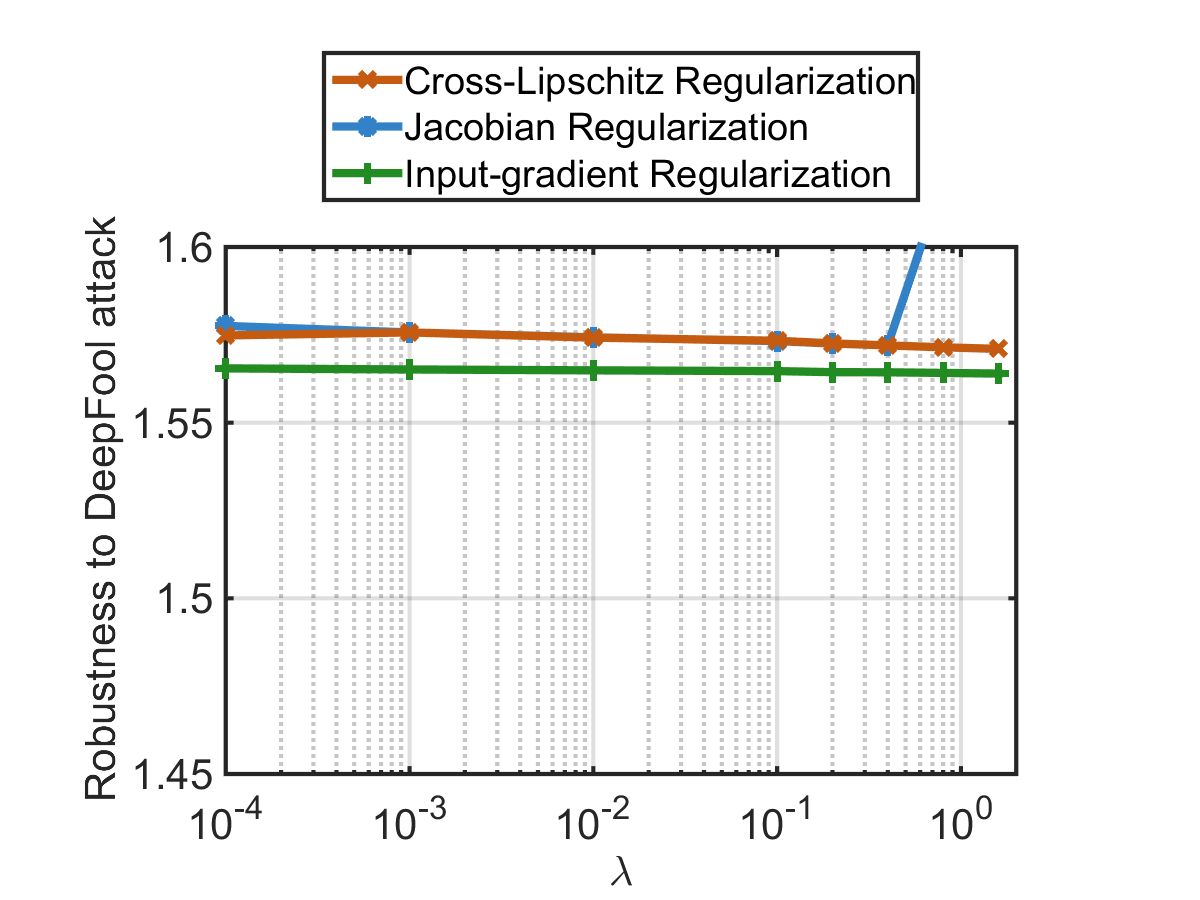}	
		\caption{}	
	\end{subfigure}
	\begin{subfigure}[b]{0.244\linewidth}
		\includegraphics[trim={0.0in 0.02in 0.5in 0.23in},clip, width=\linewidth]{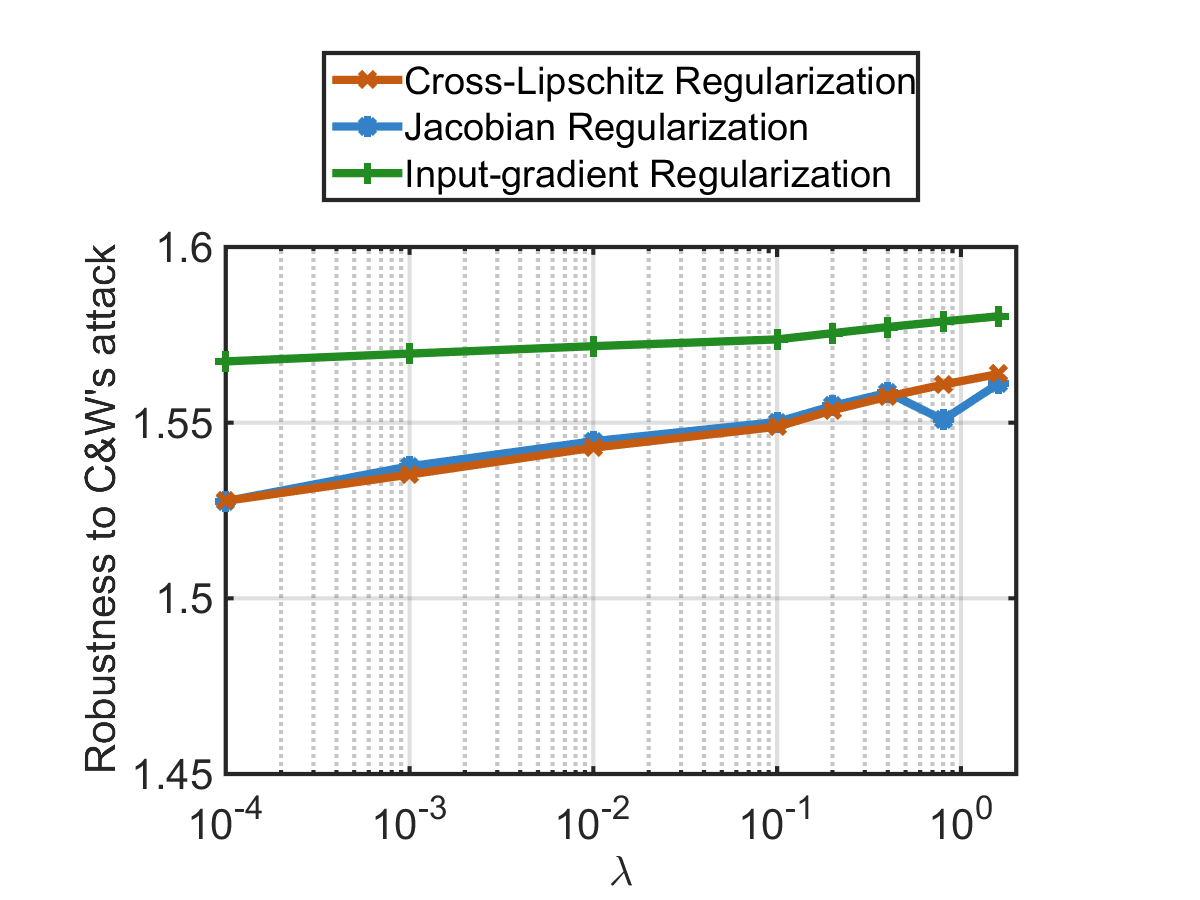}
		\caption{}
	\end{subfigure}\vskip -0.1in
	\caption{The robustness of obtained multi-class classification models evaluated with FGSM, PGD, DeepFool, and the C\&W's attacks: (a)-(d) for LeNet-300-100 and (e)-(h) for LeNet-5. Ten runs from different initializations were performed and the average results over the multiple runs are reported. The curvature regularization is not compared as approximations seem inevitable in its multi-class implementation. 
	}\vskip -0.1in
\label{fig:5} %
\end{figure*}

\section{Multi-class Classification}\label{sec:mc}

This section focuses on multi-class classification tasks. 
The notations are mostly the same as those in the binary classification. 
Suppose there are $K$ possible labels for an instance, \ie, $y\in\{0,\ldots, K-1\}$ and $K\geq2$, then for the discussed general ReLU networks, we have $n_d=K$.
Similarly, there exists a polytope $Q({\mathbf x})$ to which the input instance $\mathbf x$ belongs and on which the network $g(\cdot)$ is linear, \ie
\begin{equation}\label{eq:14}
\left.g(\mathbf x')\right|_{\mathbf x' \in Q({\mathbf x})} = V^T\mathbf x', 
\end{equation}
in which $V=[\mathbf v_{0},\ldots,\mathbf v_{K-1}]$ is a matrix with its $j$-th column $\mathbf v_j:= W_1 D_1(\mathbf x)\ldots W_{d-1} D_{d-1}(\mathbf x)\mathbf w_j$. 
For the properties of DNNs that are considered in the regularization strategies, we have the following lemma.

\begin{lemma}\textnormal{(Simplified expressions for $J$, $\nabla$, and $H$ in multi-class classification).}\label{lem:3}
Given an input instance paired with the one-hot representation of its label $(\mathbf x,\mathbf y)$, we have for $J$, $\nabla$, and $H$:
\begin{equation}\label{eq:15}
\begin{aligned}
J & = V, \\
\nabla &=V (p(\mathbf x)-\mathbf y), \\
H&=V\left(\mathrm{diag}(p(\mathbf x)) - p(\mathbf x)^T p(\mathbf x) 
\right)V^T  \\
&=\textstyle \sum_{i<j} p(\mathbf x)_i p(\mathbf x)_j(\mathbf v_i-\mathbf v_j)(\mathbf v_i-\mathbf v_j)^T.
\end{aligned}
\end{equation}
\end{lemma} \vskip -0.03in
The above expressions for $\nabla$ and $H$ seem more complex and different from those given in Lemma~\ref{lem:2}.
In particular, the local cross-Lipschitz constant seems absent in $\|\nabla\|_2$ (for the input-gradient regularizer).  
Furthermore, on account of the difficulty of decomposing the Hessian matrix $H\in\mathbb R^{n\times n}$, one might not have an analytic expression for its spectral norm and $\|\mathbf r^\ast\|_2$, in which the concerned adversarial perturbation $\mathbf r^\ast\in\mathbb R^n$ is defined similarly as in binary classification in Section~\ref{sec:binary_rob}, involving a threshold value of $\log(K)$ rather than $\log(2)$. 
To give insights as in the binary classification scenarios, we first derive bounds for the magnitude of the adversarial perturbation $\mathbf r^\ast$. 

\begin{proposition}\textnormal{(Lower bounds of $\|\mathbf r^*\|_2$ in multi-class classification).}\label{pro:5}
For the multi-class classifier with a locally linear $g(\cdot)$ and a correctly classified instance $\mathbf x$, we have the bounds
\begin{equation}\label{eq:16}
\begin{aligned}
    \|\mathbf r^*\|_2 \geq &\frac{4\|V\|_F}{K(K-1)p(\mathbf x)_y\nu^2} \\
&\quad \times\left(\sqrt{1+\frac{K(K-1)p(\mathbf x)_y \nu^2\xi}{4	(1-p(\mathbf x)_y)\|V\|^2_F}}-1\right),
\end{aligned}
\end{equation} 
and 
\begin{equation}\label{eq:17}
    \|\mathbf r^*\|_2 \geq  \frac{1}{p(\mathbf x)_y \left\| V \right\|_F}\left(\sqrt{1+\frac{p(\mathbf x)_y \xi}{(1-p(\mathbf x)_y)}}-1\right).
\end{equation}
\end{proposition}

Considering that the value of $\xi=\log(K)-\log(p(\mathbf x)_y)$ is determinate w.r.t. the prediction probability $p(\mathbf x)_y$, we can conclude from Eq.~(\ref{eq:17}) that the essential ingredients of such a lower bound are $p(\mathbf x)_y$ and $\|V\|_F$ (\ie, the Frobenius norm of an $n\times K$ matrix $V$). 
Likewise, we can easily verify that $\|V\|_F$ is a local Lipschitz constant of $g(\cdot)$. 
Somewhat unsurprisingly, a property considered in the cross-Lipschitz regularization defined as $\nu^2:=\sum\|\mathbf v_i-\mathbf v_j\|^2_2/K^2$~\cite{Hein2017} is involved in the other derived lower bound as given in Eq.~(\ref{eq:16}).
The results show that the local (cross-)Lipschitz constants and prediction probability are possibly still the essential ingredients of $\|\mathbf r^\ast\|_2$.
Apart from Proposition~\ref{pro:5}, we further know that the chained inequality $\|\nabla\|_2/2\leq\|H\|_2\leq\|V\|_F/2$ holds  by derivations from Lemma~\ref{lem:3}.  
More discussions similar to those made for binary classification in Section~\ref{sec:bcreg} will be given in Appendix E (right after the proof). 

As in binary classification, we aim to study possible connections between regularizations penalizing a squared local Lipschitz constant $\mu^2:=\|V\|^2_F$ and $\nu^2$.
Experimental results are given to show a vague equivalence.
The same MLP and convolutional architectures were adopted.
Similar to the binary classification experiments, we trained multiple baseline models for each considered architecture and fine-tuned them using different regularizations.
The same training and test policies were also kept.
We report the average results of obtained model robustness to FGSM, PGD, Deepfool, and the C\&W's attack in Figure~\ref{fig:5}.
It can be seen that the Jacobian regularization and cross-Lipschitz regularization still perform similarly across all tested $\lambda$ values, except for the ones being too large to keep the models numerically stable. 
NaN was produced in Jacobian regularized LeNet-5 if $\lambda$ was further enlarged.

\section{Conclusions}\label{sec:con}
This paper aims at exploring and analyzing possible connections between recent network-property-based regularizations for improving the adversarial robustness of DNNs. 
While the empirical effectiveness of appropriate regularizations has been demonstrated in prior arts~\cite{Hein2017,Ross2018,Jakubovitz2018,Moosavi2019}, there still lacks systematic understanding of their intrinsic functionality and connections. 
We made some comparative analyses among these regularizations and our achievements include:
\begin{description}[leftmargin=!,labelwidth=\widthof{\bfseries $\bullet\quad$},font=$\bullet$]
\item [\quad] We have analyzed regularizations on DNNs with ReLU activations from a theoretical perspective.
\item [\quad] We have presented analytic expressions for the $l_2$ and $l_\infty$ magnitudes of some approximately-optimal adversarial perturbations, and we have shown that the local cross-Lipschitz constants and prediction probability are their essential ingredients in binary classification.  
\item [\quad] We have demonstrated that, the regularizations suggest either small Lipschitz constants or small cross-Lipschitz constants, and regularizing them can be equivalent. Yet, critical discrepancies still exist between specific regularizations, mostly in handling the prediction probability.
\item [\quad] We have verified that curvature regularization~\cite{Moosavi2019} concerned in a very recent paper shows the most promising performance, and we have extended some of our analyses to multi-class classification and verified our findings with experiments.
\end{description}


\bibliographystyle{IEEEtran}
\bibliography{ref}

\begin{thebibliography}{10}
\providecommand{\url}[1]{#1}
\csname url@samestyle\endcsname
\providecommand{\newblock}{\relax}
\providecommand{\bibinfo}[2]{#2}
\providecommand{\BIBentrySTDinterwordspacing}{\spaceskip=0pt\relax}
\providecommand{\BIBentryALTinterwordstretchfactor}{4}
\providecommand{\BIBentryALTinterwordspacing}{\spaceskip=\fontdimen2\font plus
\BIBentryALTinterwordstretchfactor\fontdimen3\font minus
  \fontdimen4\font\relax}
\providecommand{\BIBforeignlanguage}[2]{{%
\expandafter\ifx\csname l@#1\endcsname\relax
\typeout{** WARNING: IEEEtran.bst: No hyphenation pattern has been}%
\typeout{** loaded for the language `#1'. Using the pattern for}%
\typeout{** the default language instead.}%
\else
\language=\csname l@#1\endcsname
\fi
#2}}
\providecommand{\BIBdecl}{\relax}
\BIBdecl

\bibitem{Szegedy2014}
C.~Szegedy, W.~Zaremba, I.~Sutskever, J.~Bruna, D.~Erhan, I.~Goodfellow, and
  R.~Fergus, ``Intriguing properties of neural networks,'' in \emph{ICLR},
  2014.

\bibitem{Goodfellow2015}
I.~J. Goodfellow, J.~Shlens, and C.~Szegedy, ``Explaining and harnessing
  adversarial examples,'' in \emph{ICLR}, 2015.

\bibitem{Madry2018}
A.~Madry, A.~Makelov, L.~Schmidt, D.~Tsipras, and A.~Vladu, ``Towards deep
  learning models resistant to adversarial attacks,'' in \emph{ICLR}, 2018.

\bibitem{Tramer2018}
F.~Tram{\`e}r, A.~Kurakin, N.~Papernot, D.~Boneh, and P.~McDaniel, ``Ensemble
  adversarial training: Attacks and defenses,'' in \emph{ICLR}, 2018.

\bibitem{Kurakin2017}
A.~Kurakin, I.~Goodfellow, and S.~Bengio, ``Adversarial machine learning at
  scale,'' in \emph{ICLR}, 2017.

\bibitem{Moosavi2019}
S.-M. Moosavi-Dezfooli, A.~Fawzi, J.~Uesato, and P.~Frossard, ``Robustness via
  curvature regularization, and vice versa,'' in \emph{CVPR}, 2019.

\bibitem{Krogh1992}
A.~Krogh and J.~A. Hertz, ``A simple weight decay can improve generalization,''
  in \emph{NeurIPS}, 1992.

\bibitem{Srivastava2014}
N.~Srivastava, G.~Hinton, A.~Krizhevsky, I.~Sutskever, and R.~Salakhutdinov,
  ``Dropout: a simple way to prevent neural networks from overfitting,''
  \emph{The Journal of Machine Learning Research}, vol.~15, no.~1, pp.
  1929--1958, 2014.

\bibitem{Cisse2017}
M.~Cisse, P.~Bojanowski, E.~Grave, Y.~Dauphin, and N.~Usunier, ``Parseval
  networks: Improving robustness to adversarial examples,'' in \emph{ICML},
  2017.

\bibitem{Hein2017}
M.~Hein and M.~Andriushchenko, ``Formal guarantees on the robustness of a
  classifier against adversarial manipulation,'' in \emph{NeurIPS}, 2017.

\bibitem{Ross2018}
A.~S. Ross and F.~Doshi-Velez, ``Improving the adversarial robustness and
  interpretability of deep neural networks by regularizing their input
  gradients,'' in \emph{AAAI}, 2018.

\bibitem{Jakubovitz2018}
D.~Jakubovitz and R.~Giryes, ``Improving dnn robustness to adversarial attacks
  using jacobian regularization,'' in \emph{ECCV}, 2018.

\bibitem{Lyu2015}
C.~Lyu, K.~Huang, and H.-N. Liang, ``A unified gradient regularization family
  for adversarial examples,'' in \emph{ICDM}, 2015.

\bibitem{Sokolic2017}
J.~Sokoli{\'c}, R.~Giryes, G.~Sapiro, and M.~R. Rodrigues, ``Robust large
  margin deep neural networks,'' \emph{IEEE Transactions on Signal Processing},
  vol.~65, no.~16, pp. 4265--4280, 2017.

\bibitem{Simon2018}
C.-J. Simon-Gabriel, Y.~Ollivier, L.~Bottou, B.~Sch{\"o}lkopf, and
  D.~Lopez-Paz, ``Adversarial vulnerability of neural networks increases with
  input dimension,'' in \emph{ICML}, 2019.

\bibitem{Papernot2017}
N.~Papernot, P.~McDaniel, I.~Goodfellow, S.~Jha, Z.~B. Celik, and A.~Swami,
  ``Practical black-box attacks against machine learning,'' in
  \emph{Proceedings of the Asia Conference on Computer and Communications
  Security}, 2017.

\bibitem{Chen2017}
P.-Y. Chen, H.~Zhang, Y.~Sharma, J.~Yi, and C.-J. Hsieh, ``Zoo: Zeroth order
  optimization based black-box attacks to deep neural networks without training
  substitute models,'' in \emph{Proceedings of the 10th ACM Workshop on
  Artificial Intelligence and Security}.\hskip 1em plus 0.5em minus 0.4em\relax
  ACM, 2017, pp. 15--26.

\bibitem{Carlini2017}
N.~Carlini and D.~Wagner, ``Towards evaluating the robustness of neural
  networks,'' in \emph{Proceedings of the IEEE Symposium on Security and
  Privacy}, 2017.

\bibitem{Papernot2016}
N.~Papernot, P.~McDaniel, S.~Jha, M.~Fredrikson, Z.~B. Celik, and A.~Swami,
  ``The limitations of deep learning in adversarial settings,'' in
  \emph{Proceedings of the IEEE European Symposium on Security and Privacy},
  2016.

\bibitem{Moosavi2016}
S.-M. Moosavi-Dezfooli, A.~Fawzi, and P.~Frossard, ``Deep{F}ool: a simple and
  accurate method to fool deep neural networks,'' in \emph{CVPR}, 2016.

\bibitem{Chen2018}
P.-Y. Chen, Y.~Sharma, H.~Zhang, J.~Yi, and C.-J. Hsieh, ``Ead: elastic-net
  attacks to deep neural networks via adversarial examples,'' in \emph{AAAI},
  2018.

\bibitem{Athalye2018}
A.~Athalye, N.~Carlini, and D.~Wagner, ``Obfuscated gradients give a false
  sense of security: Circumventing defenses to adversarial examples,'' in
  \emph{ICML}, 2018.

\bibitem{Guo2018}
Y.~Guo, C.~Zhang, C.~Zhang, and Y.~Chen, ``Sparse dnns with improved
  adversarial robustness,'' in \emph{NeurIPS}, 2018.

\bibitem{Drucker1991}
H.~Drucker and Y.~LeCun, ``Double backpropagation increasing generalization
  performance,'' in \emph{IJCNN}, 1991.

\bibitem{Nair2010}
V.~Nair and G.~E. Hinton, ``Rectified linear units improve restricted boltzmann
  machines,'' in \emph{ICML}, 2010.

\bibitem{He2015}
K.~He, X.~Zhang, S.~Ren, and J.~Sun, ``Delving deep into rectifiers: Surpassing
  human-level performance on imagenet classification,'' in \emph{CVPR}, 2015.

\bibitem{Shang2016}
W.~Shang, K.~Sohn, D.~Almeida, and H.~Lee, ``Understanding and improving
  convolutional neural networks via concatenated rectified linear units,'' in
  \emph{ICML}, 2016.

\bibitem{He2016}
K.~He, X.~Zhang, S.~Ren, and J.~Sun, ``Deep residual learning for image
  recognition,'' in \emph{CVPR}, 2016.

\bibitem{Bahdanau2015}
D.~Bahdanau, K.~Cho, and Y.~Bengio, ``Neural machine translation by jointly
  learning to align and translate,'' in \emph{ICLR}, 2015.

\bibitem{Tsipras2019}
D.~Tsipras, S.~Santurkar, L.~Engstrom, A.~Turner, and A.~Madry, ``Robustness
  may be at odds with accuracy,'' in \emph{ICLR}, 2019.

\bibitem{Lecun1998}
Y.~LeCun, L.~Bottou, Y.~Bengio, P.~Haffner \emph{et~al.}, ``Gradient-based
  learning applied to document recognition,'' \emph{Proceedings of the IEEE},
  vol.~86, no.~11, pp. 2278--2324, 1998.

\end{thebibliography}



\section*{Supplementary material}

# On Connections between Regularizations for Improving DNN Robustness **Appendices**
Yiwen Guo, Long Chen, Yurong Chen, and Changshui Zhang, *Fellow, IEEE*

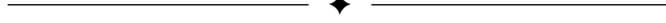

## APPENDIX A
## PROOF OF LEMMA 3.2

*Proof.* According to the definition, it is self-evident that $J = V$. As for the input-gradient, we have

$$\begin{aligned}\nabla_{\mathbf{x}}\mathcal{L}(\mathbf{x}, y) &= -V(\mathbf{y} - p(\mathbf{x})) \\ &= -y(1 - p(\mathbf{x})_y)(\mathbf{v}_+ - \mathbf{v}_-)^T,\end{aligned} \quad (18)$$

according to the chain rule, in which $y \in \{\pm 1\}$ and $\mathbf{y}$ is its one-hot vector representation. Similarly, the Hessian matrix of $\mathcal{L}(\cdot, \cdot)$ w.r.t. $\mathbf{x}$ is

$$\begin{aligned}H &= \nabla_{\mathbf{x}}\left(V(p(\mathbf{x}) - \mathbf{y})\right) \\ &= V\left(\nabla_{\mathbf{x}}\left(p(\mathbf{x}) - \mathbf{y}\right)\right)^T \\ &= V(\text{diag}(p(\mathbf{x})) - p(\mathbf{x})p(\mathbf{x})^T)V^T \\ &= p(\mathbf{x})_+ p(\mathbf{x})_- (\mathbf{v}_+ - \mathbf{v}_-)(\mathbf{v}_+ - \mathbf{v}_-)^T.\end{aligned} \quad (19)$$

□

## APPENDIX B
## PROOF OF PROPOSITION 3.1

*Proof.* From the expression of $H$ shown in Eq. (19), we know that $H$ is a rank-1 positive semi-definite matrix and its only eigenvalue becomes the maximal eigenvalue (i.e, the spectral norm). Since the instance is correctly classified, we have $p(\mathbf{x})_y > 0.5$ and $\|\mathbf{v}_+ - \mathbf{v}_-\|_2 \neq 0$ hold. Suppose that $\mathbf{x}$ and $\mathbf{v}_+ - \mathbf{v}_-$ have finite magnitude, then we further know $p(\mathbf{x})_y \neq 1.0$. From

$$H\nabla = p(\mathbf{x})_y(1 - p(\mathbf{x})_y)\|\mathbf{v}_+ - \mathbf{v}_-\|_2^2 \nabla,$$

we know that $\nabla$ should be an eigenvector corresponding to $p(\mathbf{x})_y(1 - p(\mathbf{x})_y)\|\mathbf{v}_+ - \mathbf{v}_-\|_2^2 > 0$ as the eigenvalue of $H$. Hence we have $\mathbf{u} = \pm \frac{\nabla}{\|\nabla\|_2}$ and further

$$|\nabla^T \mathbf{u}| = \|\nabla\|_2.$$

- Y. Guo is with Bytedance AI Lab. E-mail: guoyiwen.ai@bytedance.com.
- L. Chen is with the Academy for Advanced Interdisciplinary Studies, Center for Data Science, Peking University, Beijing 100871, China. E-mail: xidonglc@gmail.com.
- Y. Chen is with Intel Labs China. E-mail: yurong.chen@intel.com.
- C. Zhang is with the Department of Automation, State Key Lab of Intelligence Technologies and Systems, Tsinghua National Laboratory for Information Science and Technology, Tsinghua University, Beijing 100084, China. E-mail: zcs@mail.tsinghua.edu.cn

Y. Guo and L. Chen contribute equally to this work.

In consequence, the upper bound and low bound in Lemma 3.1 are in fact the same at this point, and it follows that

$$\begin{aligned}\|\mathbf{r}^*\|_2 &= \frac{\|\nabla\|_2}{\|H\|_2}\left(\sqrt{1 + \frac{2\|H\|_2 \xi}{\|\nabla\|_2^2}} - 1\right) \\ &= \frac{(1 - p(\mathbf{x})_y)\|\mathbf{v}_+ - \mathbf{v}_-\|_2}{p(\mathbf{x})_y(1 - p(\mathbf{x})_y)\|\mathbf{v}_+ - \mathbf{v}_-\|_2^2} \\ &\quad \left(\sqrt{1 + \frac{2p(\mathbf{x})_y(1 - p(\mathbf{x})_y)\|\mathbf{v}_+ - \mathbf{v}_-\|_2^2 \xi}{(1 - p(\mathbf{x})_y)^2 \|\mathbf{v}_+ - \mathbf{v}_-\|_2^2}} - 1\right) \\ &= \frac{1}{p(\mathbf{x})_y \|\mathbf{v}_+ - \mathbf{v}_-\|_2} \\ &\quad \left(\sqrt{1 + \frac{2p(\mathbf{x})_y \xi}{1 - p(\mathbf{x})_y}} - 1\right).\end{aligned} \quad (20)$$

□

## APPENDIX C
## PROOF OF PROPOSITION 3.2 AND 3.3

We first provide our proof of Proposition 3.2 which shows an analytic expression for $\|\hat{\mathbf{r}}^*\|_\infty$ as below.

*Proof.* According to the definition, we have:

$$\|\tilde{\mathbf{r}}^*\|_\infty := \min \|\mathbf{r}\|_\infty \quad \text{s.t.} \quad \mathcal{L}(\mathbf{x}, y) + \nabla^T \mathbf{r} + \mathbf{r}^T H \mathbf{r}/2 \geq \beta. \quad (21)$$

By substituting $H$ with its expression given in Lemma 3.2, the above constraint can be written as

$$-\xi + \nabla^T \mathbf{r} + \frac{p(\mathbf{x})_y}{2(1 - p(\mathbf{x})_y)}\left(\nabla^T \mathbf{r}\right)^2 \geq 0. \quad (22)$$

Now that $\nabla^T \mathbf{r}$ is a scalar, we can consider (22) as a quadratic inequality and equivalently we have

$$\nabla^T \mathbf{r} \geq \frac{1 - p(\mathbf{x})_y}{p(\mathbf{x})}\left(\sqrt{1 + \frac{2p(\mathbf{x})_y \xi}{1 - p(\mathbf{x})_y}} - 1\right) \quad \text{or}$$

$$\nabla^T \mathbf{r} \leq -\frac{1 - p(\mathbf{x})_y}{p(\mathbf{x})}\left(\sqrt{1 + \frac{2p(\mathbf{x})_y \xi}{1 - p(\mathbf{x})_y}} + 1\right).$$

Since it holds that $\|\mathbf{r}\|_\infty \geq \frac{\nabla^T \mathbf{r}}{\|\nabla\|_1} \geq -\|\mathbf{r}\|_\infty$ for any $\mathbf{r} \in \mathbb{R}^n$ and the equalities should be attained at $\mathbf{r} = \|\mathbf{r}\|_\infty \text{sign}(\nabla)$ and $\mathbf{r} = -\|\mathbf{r}\|_\infty \text{sign}(\nabla)$ respectively, we have

$$\|\tilde{\mathbf{r}}^*\|_\infty = \frac{1 - p(\mathbf{x})_y}{p(\mathbf{x})\|\nabla\|_1} \left( \sqrt{1 + \frac{2p(\mathbf{x})_y \xi}{1 - p(\mathbf{x})_y}} - 1 \right)$$
$$= \frac{1}{p(\mathbf{x})\|\mathbf{v}_+ - \mathbf{v}_-\|_1} \left( \sqrt{1 + \frac{2p(\mathbf{x})_y \xi}{1 - p(\mathbf{x})_y}} - 1 \right). \quad (23)$$

□

Then it is the proof of Proposition 3.3.

*Proof.* We aim at analyzing $\eta^* := \max_{\|\mathbf{r}\|_\infty \leq \epsilon} \mathcal{L}(\mathbf{x}, y) + \nabla^T \mathbf{r} + \mathbf{r}^T H \mathbf{r}/2$. According to the Hölder's inequality, we have for all $\mathbf{r} \in \mathbb{R}^n$ satisfying $\|\mathbf{r}\|_\infty \leq \epsilon$, it holds that

$$\nabla^T \mathbf{r} + \mathbf{r}^T H \mathbf{r}/2 = \nabla^T \mathbf{r} + \frac{p(\mathbf{x})_y}{2(1 - p(\mathbf{x})_y)} \left( \nabla^T \mathbf{r} \right)^2$$
$$\leq \|\mathbf{r}\|_\infty \|\nabla\|_1 + \frac{p(\mathbf{x})_y}{2(1 - p(\mathbf{x})_y)} \left( \|\mathbf{r}\|_\infty \|\nabla\|_1 \right)^2$$
$$= \epsilon \|\nabla\|_1 + \frac{p(\mathbf{x})_y}{2(1 - p(\mathbf{x})_y)} \left( \epsilon \|\nabla\|_1 \right)^2. \quad (24)$$

By further substituting the vector $\nabla$ with the expression given in Lemma 3.2, we have

$$\nabla^T \mathbf{r} + \mathbf{r}^T H \mathbf{r}/2 \leq \epsilon (1 - p(\mathbf{x})_y) \|\mathbf{v}_+ - \mathbf{v}_-\|$$
$$+ \frac{1}{2} \epsilon^2 p(\mathbf{x})_y (1 - p(\mathbf{x})_y) \|\mathbf{v}_+ - \mathbf{v}_-\|^2 \quad (25)$$

to complete our proof, and we know that the equality should be attained at $\mathbf{r} = \epsilon \cdot \text{sign}(\nabla)$. □

## APPENDIX D
## PROOF OF PROPOSITION 3.4

*Proof.* Let us denote by $\mathcal{L}'(\cdot)$ and $\mathcal{L}''(\cdot)$ the loss functions for training regularized by the Jacobian and cross-Lipschitz strategies, respectively. That is, $\mathcal{L}'(V) = -E[\log p(\mathbf{x}; V)_y] + \lambda' \mu^2$ and $\mathcal{L}''(V) = -E[\log p(\mathbf{x}; V)_y] + \lambda'' \nu^2$, in which $E[\cdot]$ calculates the sample mean rather than the population mean. It is easy to verify that $\mathcal{L}'(V)$ is strongly convex w.r.t. $V$ for single layer perceptrons and piecewise linear DNNs in which only the final layer is to be optimized, thus we have a unique optimal solution to $\min_V \mathcal{L}'(V)$. Let us denote the optimal solution as $V' = [\mathbf{v}'_+, \mathbf{v}'_-]$.

For binary classification in which $y \in \{+1, -1\}$, it holds for any matrix $V = [\mathbf{v}_+, \mathbf{v}_-]$ that,

$$-E[\log p(\mathbf{x}; V)_y]$$
$$= -E \left[ \frac{1 + y}{2} \log p(\mathbf{x}; V)_{+1} + \frac{1 - y}{2} \log p(\mathbf{x}; V)_{-1} \right]$$
$$= -E \left[ \frac{1 + y}{2} \log p(\mathbf{x}; V)_{+1} + \frac{1 - y}{2} \log (1 - p(\mathbf{x}; V)_{+1}) \right]. \quad (26)$$

By definition of the cross-entropy function, we have

$$p(\mathbf{x}; V)_{+1} = \frac{\exp(\langle \mathbf{v}_+, \mathbf{x} \rangle)}{\exp(\langle \mathbf{v}_+, \mathbf{x} \rangle) + \exp(\langle \mathbf{v}_-, \mathbf{x} \rangle)}$$
$$= \frac{\exp(\langle \mathbf{v}_+ - \mathbf{v}_-, \mathbf{x} \rangle)}{\exp(\langle \mathbf{v}_+ - \mathbf{v}_-, \mathbf{x} \rangle) + 1}, \quad (27)$$

thus further we can write $-E[\log p(\mathbf{x}; V)_y] = h(\mathbf{v}_+ - \mathbf{v}_-)$, in which $h(\cdot) : \mathbb{R}^n \to \mathbb{R}$ can also easily be verified as a convex function. We rewrite the loss function of a Jacobian-regularized training as

$$\mathcal{L}'(V) = h(\mathbf{v}_+ - \mathbf{v}_-) + \frac{1}{2}\lambda' \nu_2^2 + \frac{1}{2}\lambda' \|\mathbf{v}_+ + \mathbf{v}_-\|_2^2, \quad (28)$$

considering $2\mu^2 = \nu_2^2 + \|\mathbf{v}_+ + \mathbf{v}_-\|_2^2$. Apparently, the first two terms on the right hand side of the above equation should remain unchanged if the vector $\mathbf{v}_+ - \mathbf{v}_-$ does. Let us introduce $\hat{V}' = [\hat{\mathbf{v}}'_+, \hat{\mathbf{v}}'_-]$, in which $\hat{\mathbf{v}}'_+ = \frac{\mathbf{v}'_+ - \mathbf{v}'_-}{2}$ and $\hat{\mathbf{v}}'_- = -\hat{\mathbf{v}}'_+$. Now we have $\hat{\mathbf{v}}'_+ - \hat{\mathbf{v}}'_- = \mathbf{v}'_+ - \mathbf{v}'_-$ and

$$\mathcal{L}'(\hat{V}') - \mathcal{L}'(V') = \frac{1}{2}\lambda' \|\hat{\mathbf{v}}'_+ + \hat{\mathbf{v}}'_-\|_2^2 - \frac{1}{2}\lambda' \|\mathbf{v}'_+ + \mathbf{v}'_-\|_2^2$$
$$\leq -\frac{1}{2}\lambda' \|\mathbf{v}'_+ + \mathbf{v}'_-\|_2^2 \leq 0. \quad (29)$$

Recall that $V'$ is the optimal solution to $\min_V \mathcal{L}'(V)$, we also have $0 \leq \mathcal{L}'(\hat{V}') - \mathcal{L}'(V')$. Therefore, we have $\mathbf{v}'_+ + \mathbf{v}'_- = 0$ holds in order to avoid contradictions. The obtained equation eliminates the third term in Eq. (28). Thus we know, for $\lambda'' = \lambda'/2$, it holds that

$$\mathcal{L}'(V') = \mathcal{L}''(V'). \quad (30)$$

We now proceed similarly for an optimal solution to the problem $\min_V \mathcal{L}''(V)$. By writing $\mathcal{L}''(V) = h(\mathbf{v}_+ - \mathbf{v}_-) + \lambda'' \nu^2 = l(\mathbf{v}_+ + \mathbf{v}_-)$, we can verify that the newly introduced function $l(\cdot) : \mathbb{R}^n \to \mathbb{R}$ is strongly convex as well. Let us denote the optimal solution to $\min_\mathbf{w} l(\mathbf{w})$ as $\mathbf{w}''$, then by further introducing $\mathbf{v}''_+ = \mathbf{w}''/2$ and $\mathbf{v}''_- = -\hat{\mathbf{v}}''_+$, we can verify that

$$\mathcal{L}'(V'') = \mathcal{L}''(V''). \quad (31)$$

According to Eq. (30)-(31) and the definitions of the optimal solutions $V'$ and $V''$, we now have

$$\mathcal{L}'(V') \stackrel{(30)}{=} \mathcal{L}''(V') \stackrel{\text{def.}}{\geq} \mathcal{L}''(V'') \stackrel{(31)}{=} \mathcal{L}'(V'') \stackrel{\text{def.}}{\geq} \mathcal{L}'(V'), \quad (32)$$

which leads to $\mathcal{L}''(V) = \mathcal{L}''(V)$ for $V$ being equal to $V'$ or $V''$ and we have proved the proposition. □

## APPENDIX E
## PROOF OF PROPOSITION 4.1 AND MORE

*Proof.* For a $K$-class classification task with $y \in \{0, \ldots, K - 1\}$ and the cross-entropy loss chosen, we still have from Lemma 3.1 that

$$\frac{\|\nabla\|_2}{\|H\|_2} \left( \sqrt{1 + \frac{2\|H\|_2 \xi}{\|\nabla\|_2^2}} - 1 \right) \leq \|\mathbf{r}^*\|_2 \leq$$
$$\frac{|\nabla^T \mathbf{u}|}{\|H\|_2} \left( \sqrt{1 + \frac{2\|H\|_2 \xi}{|\nabla^T \mathbf{u}|^2}} - 1 \right), \quad (33)$$

in which $\xi := \log(K) - \mathcal{L}(\mathbf{x}, y)$. In order to derive insightful bounds of $\|\mathbf{r}^*\|_2$ with less entangled variables, we first analyze the involved networks properties $\|\nabla\|_2$ and $\|H\|_2$ and then take advantage of the monotonicity of the lower bound given above.

From the expressions summarized in Lemma 4.1 we know it holds that

$$\|\nabla\|_2 \leq 2(1 - p(\mathbf{x})_y) \|V\|_2 \leq 2(1 - p(\mathbf{x})_y) \|V\|_F, \quad (34)$$

and

$$\|H\|_2 \leq \sum_{i<j} p(\mathbf{x})_i p(\mathbf{x})_j \|\mathbf{v}_i - \mathbf{v}_j\|_2^2. \tag{35}$$

By utilizing two simple inequalities $\|\mathbf{v}_i - \mathbf{v}_j\|_2^2 \leq 2(\|\mathbf{v}_i\|_2^2 + \|\mathbf{v}_j\|_2^2), \forall i, j$, and $p(\mathbf{x})_i(1 - p(\mathbf{x})_i) \leq p(\mathbf{x})_y(1 - p(\mathbf{x})_y), \forall i$, we further get to know that

$$\begin{aligned} \|H\|_2 &\leq \sum_{i \neq j} p(\mathbf{x})_i p(\mathbf{x})_j \left(\|\mathbf{v}_i\|_2^2 + \|\mathbf{v}_j\|_2^2\right) \\ &\leq 2p(\mathbf{x})_y \left(1 - p(\mathbf{x})_y\right) \|V\|_F^2. \end{aligned} \tag{36}$$

The lower bound in Eq. (33) in terms of $\|H\|_2$ or $\|\nabla\|_2$ is monotonically decreasing, thus we have

$$\begin{aligned} \|\mathbf{r}^*\|_2 &\geq \frac{2(1 - p(\mathbf{x})_y) \|V\|_F}{2p(\mathbf{x})_y (1 - p(\mathbf{x})_y) \|V\|_F^2} \\ &\quad \left(\sqrt{1 + \frac{4p(\mathbf{x})_y (1 - p(\mathbf{x})_y) \|V\|_F^2 \xi}{4(1 - p(\mathbf{x})_y)^2 \|V\|_F^2}} - 1\right) \\ &\geq \frac{1}{p(\mathbf{x})_y \|V\|_F} \\ &\quad \left(\sqrt{1 + \frac{p(\mathbf{x})_y \xi}{(1 - p(\mathbf{x})_y)}} - 1\right). \end{aligned} \tag{37}$$

Now that we have derived the second bound in the proposition, we can turn to deriving the first one. Again, according to the monotonicity of the bound in Eq. (33), we get to know

$$\begin{aligned} \|\mathbf{r}^*\|_2 &\geq \frac{2(1 - p(\mathbf{x})_y) \|V\|_F}{\sum_{i<j} p(\mathbf{x})_i p(\mathbf{x})_j \|\mathbf{v}_i - \mathbf{v}_j\|_2^2} \\ &\quad \left(\sqrt{1 + \frac{2 \sum_{i<j} p(\mathbf{x})_i p(\mathbf{x})_j \|\mathbf{v}_i - \mathbf{v}_j\|_2^2 \xi}{4(1 - p(\mathbf{x})_y)^2 \|V\|_F^2}} - 1\right) \\ &\geq \frac{4\|V\|_F}{K(K-1) p(\mathbf{x})_y \nu^2} \\ &\quad \left(\sqrt{1 + \frac{K(K-1) p(\mathbf{x})_y \nu^2 \xi}{4(1 - p(\mathbf{x})_y) \|V\|_F^2}} - 1\right), \end{aligned} \tag{38}$$

in which $\nu := \sum \|\mathbf{v}_i - \mathbf{v}_j\|_2^2 / K^2$ is first introduced in the cross-Lipschitz regularization [1] and the evident inequality $p(\mathbf{x})_i p(\mathbf{x})_j \leq p(\mathbf{x})_y(1 - p(\mathbf{x})_y), \forall i, j$, is utilized. In general multi-class classification scenarios, it is challenging to compare $K(K-1)\nu^2/4$ with $\|V\|_F^2$ directly, hence we are uncertain which of the derived lower bounds would be tighter. $\square$

In addition to the above proof, we will discuss as follows more about the adversarial robustness in multi-class classification scenarios and possible connections between regularizations at this point. First of all, we try to analyze worst-case perturbations for $l_\infty$ **norm-based attacks**. We care about the lower bound of $\|\tilde{\mathbf{r}}^*\|_\infty$, in which $\tilde{\mathbf{r}}^* \in \mathbb{R}^n$ is defined similarly as in binary classification except for a different value of threshold. We notice that the inequality

$$\begin{aligned} -\xi + \|\nabla\|_1 \|\tilde{\mathbf{r}}^*\|_\infty + \frac{n\|H\|_1 \|\tilde{\mathbf{r}}^*\|_\infty^2}{2} &\geq \\ -\xi + \nabla^T \mathbf{r}^* + \frac{(\mathbf{r}^*)^T H \mathbf{r}^*}{2} &\geq 0, \end{aligned} \tag{39}$$

holds, on account of Hölder's inequality and $(\mathbf{r}^*)^T H \mathbf{r}^* \leq n\|H\|_1 \|\tilde{\mathbf{r}}^*\|_\infty^2$. Therefore, we can get

$$\|\tilde{\mathbf{r}}^*\|_\infty \geq \frac{2\xi}{\sqrt{\|\nabla\|_1^2 + 2n\|H\|_1 \xi} + \|\nabla\|_1}. \tag{40}$$

According to the definition of induced matrix norms, we additionally have $\|\nabla\|_1 \leq 2(1 - p(\mathbf{x})_y)\|V\|_1$ and $\|H\|_1 \leq 2p(\mathbf{x})_y(1 - p(\mathbf{x})_y)\|V\|_1\|V\|_\infty$ to achieve a lower bound of $\|\tilde{\mathbf{r}}^*\|_\infty$ without the Hessian and the input-gradient getting involved. Apparently, the prediction probability and matrix norms of $V$ are still essential ingredients of the DNN robustness under such circumstance, as we have presented in the multi-class $l_2$ case.

**Cross-Lipschitz and Lipschitz:** Even though it is difficult to compare $K(K-1)\nu^2/4$ with $\|V\|_F^2$ directly, we know from Proposition 4.1 that penalizing the two quantities both contribute to improving the DNN robustness, and in fact they have already been adopted in the cross-Lipschitz and Jacobian regularizations, respectively. We now discuss about their connections with other network properties and regularizations. As have been introduced in the main body of our paper, we have that the chained inequality $\|\nabla\|_2^2/2 \leq \|H\|_2 \leq \|V\|_F^2/2$ holds, which is obtained in virtue of:

$$\|H\|_2 \leq 2p(\mathbf{x})_y(1 - p(\mathbf{x})_y) \|V\|_F^2 \leq \frac{1}{2}\|V\|_F^2 \tag{41}$$

and

$$\begin{aligned} \|H\|_2 &= \|\sum p(\mathbf{x})_i (\mathbf{v}_i - \bar{\mathbf{v}})(\mathbf{v}_i - \bar{\mathbf{v}})^T\|_2 \\ &\geq p(\mathbf{x})_y \|\mathbf{v}_y - \bar{\mathbf{v}}\|_2^2 \geq p(\mathbf{x})_y \|\nabla\|_2^2 \\ &\geq \frac{1}{2}\|\nabla\|_2^2, \end{aligned} \tag{42}$$

in which $\bar{\mathbf{v}} := V p(\mathbf{x})$. Similarly, we can as well derive a chained inequality for the quantity adopted in the cross-Lipschitz regularization (and the first bound in Proposition 4.1) as $\|\nabla\|_2^2/2 \leq \|H\|_2 \leq K(K-1)\nu^2/8$, by taking advantage of Eq. (42) and:

$$\begin{aligned} \|H\|_2 &\leq \sum_{i<j} p(\mathbf{x})_i p(\mathbf{x})_j \|\mathbf{v}_i - \mathbf{v}_j\|_2^2 \\ &\leq \frac{1}{2} K(K-1) p(\mathbf{x})_y (1 - p(\mathbf{x})_y) \nu^2 \\ &\leq \frac{1}{8} K(K-1) \nu^2. \end{aligned} \tag{43}$$

# APPENDIX F
# SETTINGS AND CIFAR-10 RESULTS

In this section, we first introduce our experimental settings on MNIST, involving the training and test policies, DNN architectures, evaluation metrics, etc., and then provide results on CIFAR-10.

**Experimental Settings:** The official training/test split of MNIST [2] is utilized. As briefly introduced, we train/test with an MLP codenamed "LeNet-300-100" and a convolutional neural network codenamed "LeNet-5" on MNIST. The former is comprised of two parameterized fully-connected layers, and the latter contains two convolutional layers, two max-pooling layers and two fully-connected layers. We trained 10 models each from different initializations for them as references, and fine-tuned the obtained models with different regularizations discussed in this paper to


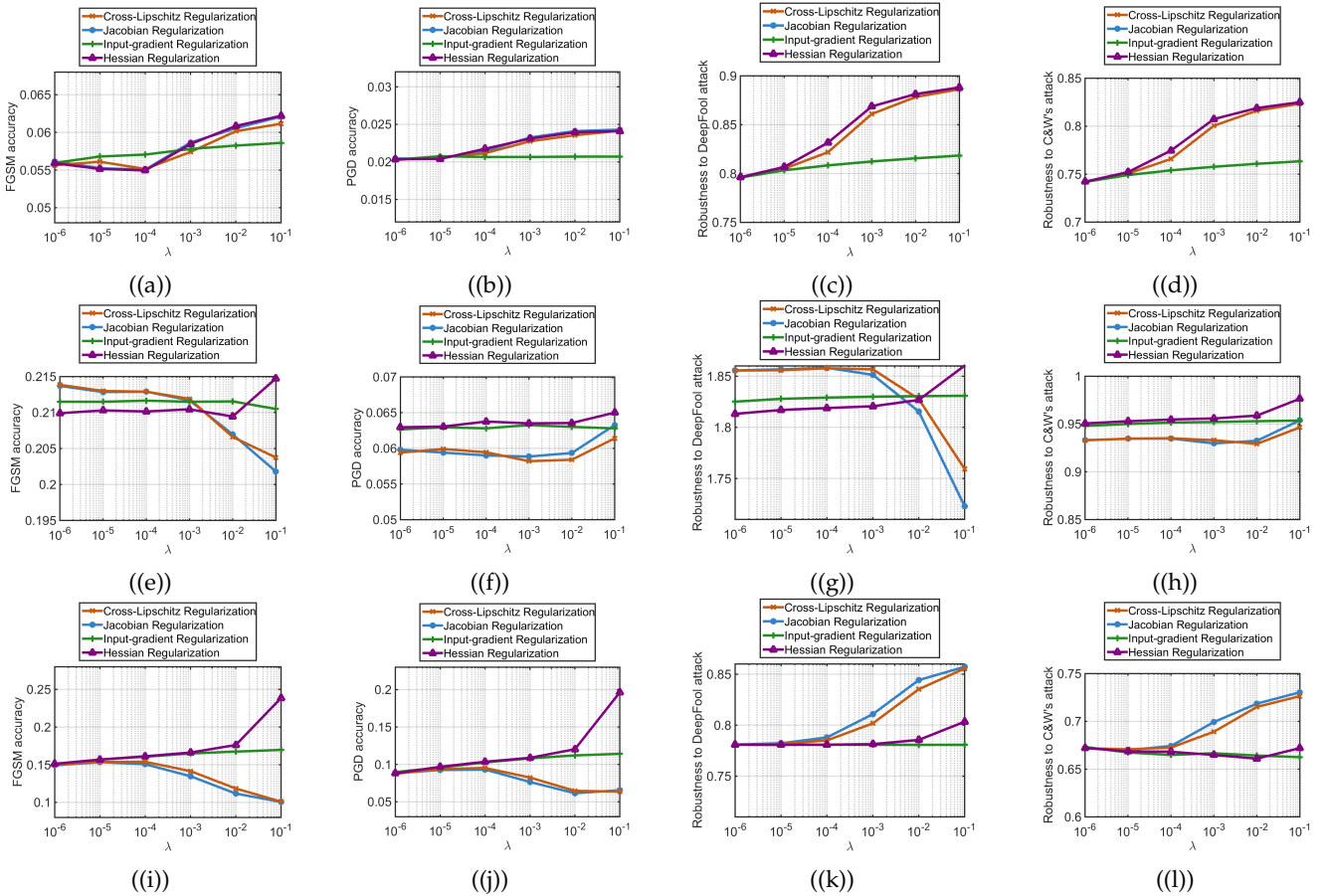

Fig. 6: The robustness of *all* obtained binary classification models evaluated with FGSM, PGD, DeepFool and the C&W's attacks on CIFAR-10: (a)-(d) for the four-layer convolutional network with batch normalizations, (e)-(h) for the VGG-like network, and (i)-(l) for ResNet. Ten runs from different initializations are performed and the average results are reported for fair comparisons.

evaluate their performance under adversarial attacks. All experiments were performed on a single NVIDIA Titan X GPU, and official implementations from the authors of the regularizations were adopted. TensorFlow [3] and Clever-Hans [4] were used.

We directly applied the training policies suggested in the Caffe model zoo [5] for LeNet-300-100 and LeNet-5, and we trained models on MNIST with a common batch size of 64 for 50,000 iterations such that they all definitely reached the plateau. For the binary LeNet-300-100 and LeNet-5 models, we achieved prediction accuracies of $99.57 \pm 0.05\%$ and $99.86 \pm 0.03\%$, respectively. To evaluate the adversarial robustness of DNN models, we chose four prevalent attacks, i.e., FGSM [6], PGD [7], DeepFool [8], and the C&W's attack [9], two of which are $l_\infty$ norm-based and the other two are $l_2$ norm-based. With $\epsilon = 0.1$, the prediction accuracies of the reference models degraded significantly (to $34.91 \pm 0.81$ and $86.32 \pm 4.86\%$) on FGSM examples and (to $7.84 \pm 0.50\%$ and $35.10 \pm 1.97\%$) on PGD adversarial examples. In multi-class scenarios, we similarly had reference models with high prediction accuracies ($98.08 \pm 0.08\%$ and $99.10 \pm 0.05$ for LeNet-300-100 and LeNet-5 respectively) on the benign test set, yet a deteriorating effect can be observed on the adversarial examples.

For training on MNIST, we regularized with various $\lambda$ values chosen from $\{10^{-4}, 10^{-3}, 10^{-2}, 0.1, 0.2, 0.4, 0.8, 1.6\}$. Such a set should cover many suggested values for setting this hyper-parameter in the literature, and we also noticed in the experiments that further enlarging the value of $\lambda$ would probably cause numerical instability during training and generated NaN in the network gradients. In fact, with a $\lambda$ as large as 1.6 on the tested dataset, most of the terms aimed to be penalized have become extremely small (typically with an order of magnitude $\leq 10$) on the training instances, thus more attention should be paid to their numerical ranges and stability. Curvature regularization, however, was stable over the range of our tested $\lambda$, and we also observed that its robustness to the C&W's attack increased to nearly 2.0 (i.e., the required magnitude of perturbations to successfully fool the model was ~2.0 in average) with $\lambda = 6.4$.

**Experimental results on CIFAR-10:** As introduced, we also performed experiments on CIFAR-10, with some much deeper networks than those tested on MNIST. To be concrete, we chose a four-layer convolutional network similar to LeNet-5 but with batch normalization [10] incorporated, a VGG-like network [11] (incorporating twelve convolutional layers and two fully-connected layers) and a ResNet [12] (incorporating 31 convolutional layers, a single fully-connected layers and no batch normalization). Again, we trained 10 models each from different initializations for them as references, and fine-tuned the obtained models with different regularizations. We trained them for 100,000 iterations to

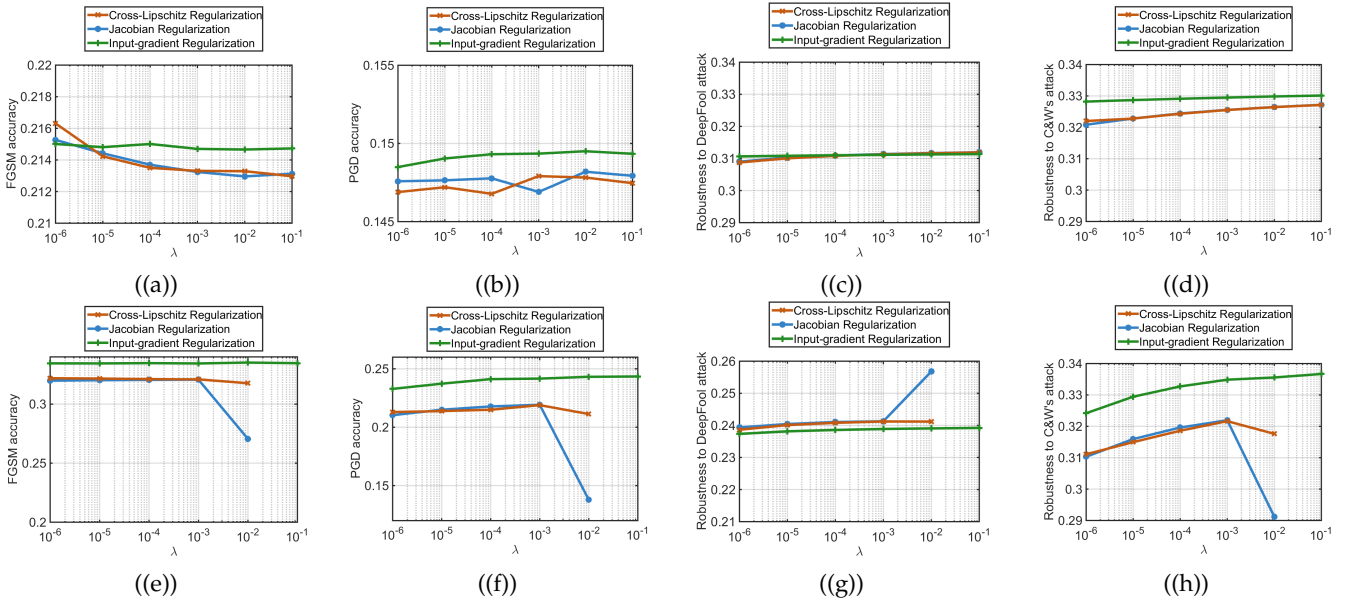

Fig. 7: The robustness of obtained multi-class classification models evaluated with FGSM, PGD, DeepFool, and the C&W's attacks: (a)-(d) for the four-layer convolutional network incorporating batch normalizations and (e)-(h) for the VGG-like DNN. The curvature regularization was not compared as approximations seem inevitable in its multi-class implementation. NaN is triggered with $\lambda = 0.1$ on Jacobian and cross-Lipschitz regularized VGG-like models. ResNets are not evaluated due to the limlited computational resources.

ensure convergence, and we decayed the learning rate by 10-fold at iterations 60,000, and 80,000.

We report the robustness of regularized models to adversarial attacks in Figure 6. The hyper-parameter $\epsilon$ for $l_\infty$ attacks was set to be 0.05, and $\lambda$ was chosen to be smaller (than the values on MNIST) to guarantee a stable training process, from $\{10^{-6}, 10^{-5}, 10^{-4}, 10^{-3}, 10^{-2}, 0.1\}$. We observed that even with the hyper-parameter as small as 0.1, it is possible to produce NaN during the training of some multi-class models on CIFAR-10, but for the binary classification models, further increasing $\lambda$ might lead to even stronger adversarial robustness. Nevertheless, performing a grid search of the best $\lambda$ for each model is beyond the scope of this paper, so we might not achieve the optimal performance of the discussed methods in Figure 6. Multi-class results are provided in Figure 7.

## APPENDIX G
## THE LOGISTIC LOSS AND OPEN QUESTIONS

It is mentioned that our theoretical results in binary classification with cross-entropy loss generalize to the case trained with **logistic loss**, which calculates $\log(1 + \exp(-y\mathbf{v}^T\mathbf{x}))$ for an instance $\mathbf{x}$ with label $y \in \{\pm 1\}$, in which $\mathbf{v} := W_1 D_1(\mathbf{x}) \ldots W_{d-1} D_{d-1}(\mathbf{x}) \mathbf{w}_d$ is an $n$-dimensional vector. Now let us discuss more about it. Equivalently, we rewrite the logistic loss as $-\log(p(\mathbf{x})_y)$, in which $p(\mathbf{x})_- = 1/(1 + \exp(\mathbf{v}^T\mathbf{x}))$ and $p(\mathbf{x})_+ = 1 - p(\mathbf{x})_-$, hence if $V = [\mathbf{v}, \mathbf{0}]$ we know by simple derivations that $\mathbf{v}$ plays the same role (in training with the logistic loss) as the vector $\mathbf{v}_+ - \mathbf{v}_-$ (in training with the cross-entropy loss). That being said, what follows is coincident with those presented in the paper. Future work should include studies on other loss functions.

Just as also mentioned in the main body of our paper, we consider the local (cross-)Lipschitz constants and the prediction probability $p(\mathbf{x})_y$ separately in regularizers for simplicity reasons, and we have our conclusions hold even if their mutual influence is rigorously analyzed. Regarding the mutual influence in binary classification, it is mostly on account of $\nu = \|\mathbf{v}_+ - \mathbf{v}_-\|_2$. Since the prediction probabilities $p(\mathbf{x})_+$ and $p(\mathbf{x})_-$ can be written as functions w.r.t. $\nu$ and $\gamma := |\check{\mathbf{w}}^T\mathbf{x}|$, in which $\check{\mathbf{w}} := (\mathbf{v}_+ - \mathbf{v}_-)/\nu$ is a normalized vector, we cast the problem as discussing about the monotonicity of $\|\mathbf{r}^*\|_2$ in terms of $\nu$ with care. We emphasize that it is actually a bit complicated, as the monotonicity highly depend on the value of $\gamma$, and we accomplish the task by calculating the derivative of regularizers including $\|\nabla\|_2^2$ and $\|H\|_2$ w.r.t. $\nu$, respectively. Specifically, we have

$$\frac{\partial \|\nabla\|_2^2}{\partial \nu} = \frac{2\nu \left(1 - (\gamma\nu - 1)\exp(\gamma\nu)\right)}{(1 + \exp(\gamma\nu))^3} \tag{44}$$

and

$$\frac{\partial \|H\|_2}{\partial \nu} = \frac{\nu \left(2 + \gamma\nu - (\gamma\nu - 2)\exp(\gamma\nu)\right)\exp(\gamma\nu)}{(1 + \exp(\gamma\nu))^3}. \tag{45}$$

By solving $\partial \|\nabla\|_2^2/\partial \nu = 0$ and $\partial \|H\|_2/\partial \nu = 0$, we can get the critical points as $1.28/\gamma$ and $2.40/\gamma$. That being said, if $\nu\gamma \leq 1.28$ and $\nu\gamma \leq 2.40$ are fulfilled, then penalizing scaled $\|\nabla\|_2^2$ and $\|H\|_2$ indicate a smaller local Lipschitz constant $\nu$, respectively. We evaluate $\nu\gamma$ with models trained on MNIST and CIFAR-10, and we find that the conditions are satisfied for almost all training instances. More specifically, the average value of $\nu\gamma$ on the LeNet-300-100 references is only roughly 0.001 and the largest value is roughly 1.570. After training with regularizations, the values of both $\nu\gamma$ and $\nu$ become orders of magnitude smaller, making the results rigorously hold for all instances. It can be interesting to evaluate whether it is the gap between 1.28 and 2.40 that affects the regularization performance in future works. Note that similar analyses can be made to the magnitude of $\mathbf{r}^*$,

but we feel the result in that case makes less sense since the definition of $\mathbf{r}^*$ is subject to approximations.

## APPENDIX H
## MULTI-CLASS REGULARIZATIONS

We did not test the curvature regularization in multi-class scenarios, mostly because some approximations seem inevitable. With approximations, it is relatively difficult to tell whether there will be functional equivalence or not from experimental results. As have been mentioned, through the lens of our study, it is possible to develop some more regularization methods by consolidating the current ones and their essential ingredients. We would like to study it more carefully in future work and compare with the current multi-class curvature approximators if possible.

## REFERENCES

[1] M. Hein and M. Andriushchenko, "Formal guarantees on the robustness of a classifier against adversarial manipulation," in *NeurIPS*, 2017.

[2] Y. LeCun, L. Bottou, Y. Bengio, P. Haffner *et al.*, "Gradient-based learning applied to document recognition," *Proceedings of the IEEE*, vol. 86, no. 11, pp. 2278–2324, 1998.

[3] M. Abadi, A. Agarwal, P. Barham, E. Brevdo, Z. Chen, C. Citro, G. S. Corrado, A. Davis, J. Dean, M. Devin, S. Ghemawat, I. Goodfellow, A. Harp, G. Irving, M. Isard, Y. Jia, R. Jozefowicz, L. Kaiser, M. Kudlur, J. Levenberg, D. Mané, R. Monga, S. Moore, D. Murray, C. Olah, M. Schuster, J. Shlens, B. Steiner, I. Sutskever, K. Talwar, P. Tucker, V. Vanhoucke, V. Vasudevan, F. Viégas, O. Vinyals, P. Warden, M. Wattenberg, M. Wicke, Y. Yu, and X. Zheng, "TensorFlow: Large-scale machine learning on heterogeneous systems," 2015, software available from tensorflow.org. [Online]. Available: http://tensorflow.org/

[4] N. Papernot, F. Faghri, N. Carlini, I. Goodfellow, R. Feinman, A. Kurakin, C. Xie, Y. Sharma, T. Brown, A. Roy, A. Matyasko, V. Behzadan, K. Hambardzumyan, Z. Zhang, Y.-L. Juang, Z. Li, R. Sheatsley, A. Garg, J. Uesato, W. Gierke, Y. Dong, D. Berthelot, P. Hendricks, J. Rauber, and R. Long, "Technical report on the cleverhans v2.1.0 adversarial examples library," *arXiv preprint arXiv:1610.00768*, 2018.

[5] Y. Jia, E. Shelhamer, J. Donahue, S. Karayev, J. Long, R. Girshick, S. Guadarrama, and T. Darrell, "Caffe: Convolutional architecture for fast feature embedding," in *MM*, 2014.

[6] I. J. Goodfellow, J. Shlens, and C. Szegedy, "Explaining and harnessing adversarial examples," in *ICLR*, 2015.

[7] A. Madry, A. Makelov, L. Schmidt, D. Tsipras, and A. Vladu, "Towards deep learning models resistant to adversarial attacks," in *ICLR*, 2018.

[8] S.-M. Moosavi-Dezfooli, A. Fawzi, and P. Frossard, "DeepFool: a simple and accurate method to fool deep neural networks," in *CVPR*, 2016.

[9] N. Carlini and D. Wagner, "Towards evaluating the robustness of neural networks," in *Proceedings of the IEEE Symposium on Security and Privacy*, 2017.

[10] S. Ioffe and C. Szegedy, "Batch normalization: Accelerating deep network training by reducing internal covariate shift," in *ICML*, 2015.

[11] K. Neklyudov, D. Molchanov, A. Ashukha, and D. P. Vetrov, "Structured bayesian pruning via log-normal multiplicative noise," in *NeurIPS*, 2017.

[12] K. He, X. Zhang, S. Ren, and J. Sun, "Delving deep into rectifiers: Surpassing human-level performance on imagenet classification," in *CVPR*, 2015.

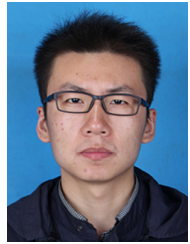

**Yiwen Guo** received the B.E. degree from Wuhan University, Wuhan, China, in 2011, and the Ph.D. degree from Tsinghua University, Beijing, China in 2016. He is a research scientist at Bytedance AI Lab, Beijing. Prior to this, he was a staff research scientist at Intel Labs China. His current research interests include computer vision, pattern recognition, and machine learning.

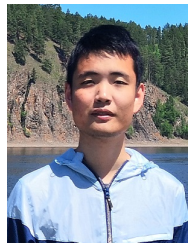

**Long Chen** received the B.S. degree in mathematics and M.S. degree in data science from Peking University, in 2016 and 2019, respectively. His current research interests include machine learning, statistics and financial data analysis.

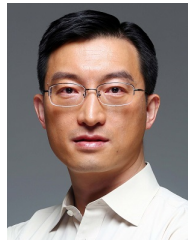

**Yurong Chen** received the B.S. and Ph.D. degrees from Tsinghua University, Beijing, China, in 1998 and 2002, respectively. He joined Intel in 2004 after completing the postdoctoral research in the Institute of Software, CAS, where he is currently a Principal Research Scientist and Director of Cognitive Computing Lab at Intel Labs China, responsible for leading visual cognition and machine learning research for Intel platforms. He received one "Intel China Award" and 3 Intel Labs Academic Awards – "Gordy Awards" for delivering leading visual analytics and understanding technologies to impact Intel platforms/solutions. He has published over 60 papers and holds over 50 issued/pending patents.

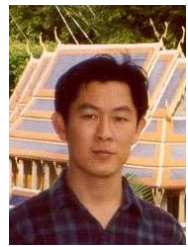

**Changshui Zhang** received the B.E. degree in mathematics from Peking University, Beijing, China, in 1986, and the M.S. and Ph.D. degrees in control science and engineering from Tsinghua University, Beijing, in 1989 and 1992, respectively. In 1992, he joined the Department of Automation, Tsinghua University, where he is currently a professor. His research interests include pattern recognition and machine learning. He is a Fellow member of the IEEE.



\end{document}